%% file: main.tex
\documentclass[10pt,twocolumn,letterpaper]{article}

\usepackage[T1]{fontenc}
\usepackage{times}

\usepackage{cvpr}              
\input{preamble}
\definecolor{cvprblue}{rgb}{0.21,0.49,0.74}
\usepackage[pagebackref,breaklinks,colorlinks,allcolors=cvprblue]{hyperref}
\usepackage{amsmath}
\usepackage{algorithmicx}
\usepackage{algpseudocode}
\usepackage{booktabs,multirow,pifont}
\usepackage{amssymb,pifont}
\usepackage{siunitx}
\usepackage{array}
\newcolumntype{M}{c}
\newcolumntype{P}{c}
\newcolumntype{T}{c}
\newcommand{\stYesYes}{\(\checkmark\!/\!\checkmark\)} 
\newcommand{\stYesNo}{\(\checkmark\!/\!\times\)}      
\newcommand{\stNone}{\ensuremath{\varnothing}}
\newcommand{\best}[1]{\textbf{#1}}
\newcommand{\secbest}[1]{\underline{#1}}

\usepackage{tabularx}
\newcolumntype{L}{>{\raggedright\arraybackslash}X} 
\newcolumntype{G}{>{\centering\arraybackslash}m{0.85cm}}
\usepackage{array}
\newcolumntype{C}[1]{>{\centering\arraybackslash}p{#1}}

\algrenewcommand\algorithmiccomment[1]{\hfill{\footnotesize // #1}}

\makeatletter
\@ifundefined{c@algorithm}{\newcounter{algorithm}}{}
\renewcommand{\thealgorithm}{\arabic{algorithm}}
\newenvironment{WideAlgo}[1]{%
  \refstepcounter{algorithm}%
  \begin{figure*}[t]
    \centering
    \begin{minipage}{0.96\textwidth}  
      \small
      \hrule height 0.6pt \vspace{2pt}
      {\bfseries Algorithm~\thealgorithm\ #1}\par
      \hrule height 0.6pt \vspace{4pt}
      \begin{algorithmic}[1]
}{%
      \end{algorithmic}
      \vspace{2pt}\hrule height 0.6pt
    \end{minipage}
  \end{figure*}
}
\makeatother

\def\paperID{***} 
\def\confName{CVPR}
\def\confYear{2026}

\title{Selective, Regularized, and Calibrated: Harnessing Vision Foundation Models for Cross-Domain Few-Shot Semantic Segmentation}

\usepackage{marvosym}

\author{
Junyuan Ma$^{1,2}$\textsuperscript{*},
Xunzhi Xiang$^{1}$,
Wenbin Li$^{1}$,
Qi Fan$^{1}$\textsuperscript{\Letter},
Yang Gao$^{1}$\\
$^{1}$Nanjing University \quad
$^{2}$University of Chinese Academy of Sciences\\
{\tt\small Project Page: \href{https://zhiyuan624.github.io/HERA-CDFSS/}{https://zhiyuan624.github.io/HERA-CDFSS/}}
}

\begin{document}
\maketitle

\begingroup
\renewcommand\thefootnote{}
\footnotetext{\textsuperscript{*} Work done during an internship at Nanjing University.}
\footnotetext{\textsuperscript{\Letter} Corresponding author.}
\endgroup

\input{sec/0_abstract}    
\input{sec/1_intro}
\input{sec/2_related}
\input{sec/3_method}
\input{sec/4_experiments}

\input{sec/5_conclusion}

\input{sec/X_suppl}

\clearpage
{
    \small
    \bibliographystyle{ieeenat_fullname}
    \bibliography{main}
}

\end{document}

%% file: sec/0_abstract.tex
\vspace{-0.4em}
\begin{abstract}
Vision foundation models (VFMs) have achieved strong performance across various vision tasks. However, it still remains challenging to apply VFMs for cross-domain few-shot segmentation (CD-FSS), which segments objects of novel classes under domain shifts using only a few labeled exemplars. The challenge is mainly driven by two factors: (1) limited labeled exemplars per novel class relative to the scale of VFM pre-training, making the model prone to overfitting during retraining, and (2) target-domain shifts underrepresented during pre-training, inducing cross-domain inconsistency and layerwise sensitivity. To address these issues, we propose Hierarchical Exemplar Representation Adaptation (HERA), a three-stage select-regularize-calibrate VFM-based segmentation framework that learns effectively from limited labels and adapts to novel domains without source-data retraining. We first design Hierarchical Layer Selection (HLS) to adaptively identify the most informative VFM layer using a data-dependent Exemplar Transfer Risk (ETR) computed for each candidate layer. Then, Prior-Guided Regularization (PGR) regularizes interactions on the selected representation, yielding well-structured local signals for the subsequent stage. Furthermore, Pixelwise Adaptive Calibration (PAC) combines the selected representation with the refined interaction maps to calibrate pixelwise predictions, producing consistent masks. Together, these stages form a hierarchical select–regularize–calibrate pipeline that guides frozen VFM features in new domains while fine-tuning less than 2.7\% of parameters at test time. Extensive experiments show that HERA surpasses the state-of-the-art by more than 4.1 mIoU across multiple CD-FSS benchmarks.
\end{abstract}
\vspace{-0.5em}

%% file: sec/1_intro.tex
\section{Introduction}
\label{sec:intro}
Few-shot semantic segmentation (FSS) aims to generate pixel-level predictions for novel classes using only a few labeled support exemplars. Existing methods~\citep{wang2019panet,min2021hypercorrelation,zhang2021few,fan2022self} learn class-agnostic correspondences from constructed pairs of support and query images, which transfers knowledge from base to novel classes and yields strong in-domain generalization. However, when the deployment distribution differs from the training distribution, the learned correspondences and class prototypes degrade, leading to large performance drops. This motivates cross-domain few-shot segmentation (CD-FSS), which seeks to generalize to novel classes in unseen target domains under distribution shift, using only a few labeled support exemplars.

\begin{figure}[tbp]
  \centering
  \includegraphics[page=1,width=\linewidth]{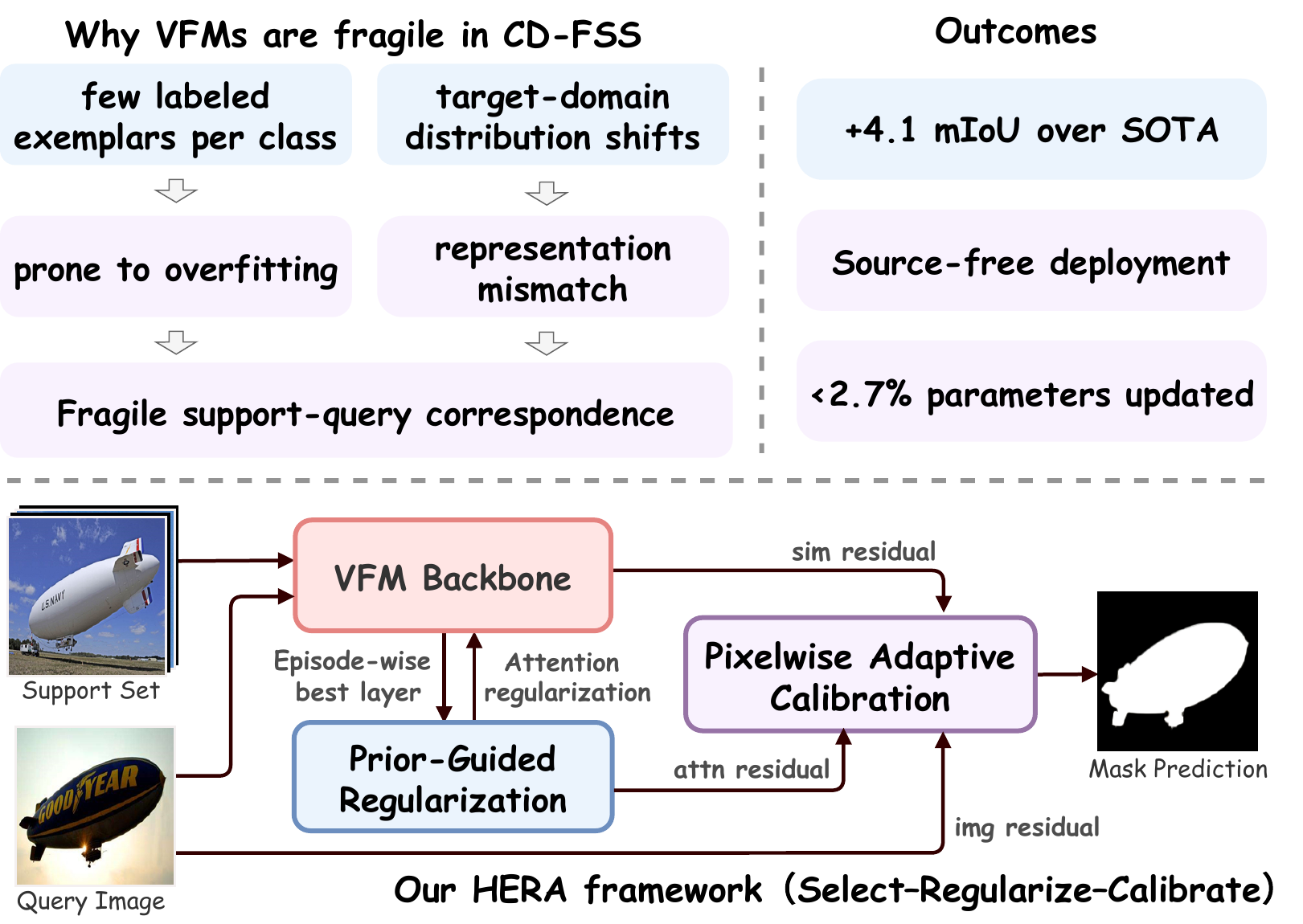}
  \vspace{-0.2in}
  \caption{\small Scarce labels and target domain shift co-occur in CD-FSS, making VFMs fragile in deployment. Retraining tends to overfit and misalign representations, thereby destabilizing support–query correspondence. Our method HERA, a three-stage select-regularize-calibrate framework, adapts a frozen VFM at test time with only limited labeled exemplars and no source data, achieving state-of-the-art performance.}
  \label{fig:introduction}
  \vspace*{-0.1in}
\end{figure}

Existing CD-FSS methods~\citep{herzog2024adapt,tong2024lightweight,nie2024cross} predominantly rely on CNN backbones, and typically extend in-domain FSS either by retraining on source data with domain-generalization techniques or by mining cross-image correspondences. Although effective in controlled settings, these approaches are costly and depend on source data. Their convolutional inductive biases limit long-range reasoning and robustness under distribution shift, leading to overfitting with sparse labels. Given these limitations, replacing CNN backbones with vision foundation models (VFMs) yields stronger and more transferable representations.

VFMs are pretrained vision backbones that provide transferable representations across recognition, segmentation, and multimodal tasks~\citep{oquab2023dinov2,chen2024internvl,simeoni2025dinov3}. Despite these strengths, applying them to CD-FSS is challenging for two reasons. First, the few-shot regime offers only a handful of labeled exemplars per novel class, so retraining is prone to overfitting, source-dependent, and computationally prohibitive for large models. Second, distribution shifts place target domains outside pretraining coverage, so using frozen features or adapting all layers indiscriminately yields cross-domain inconsistency and pronounced layer sensitivity, which destabilize correspondence and pixel predictions. Empirically, VFM features exhibit sharp layerwise variation in transferability under shift~\citep{raghu2021vision,lee2022surgical,roh2024levi}. Adapting a single stable layer reduces degrees of freedom and curbs overfitting. Notably, most recent VFM-based CD-FSS methods \citep{he2024apseg,tong2025self,liu2025devil} still assume source-domain pretraining, which directly conflicts with constraints on source-data access and is prone to overfitting in few-shot regimes. Therefore, it is necessary to adopt source-free test-time adaptation~\citep{liang2025comprehensive} that first selects the most informative layer for adaptation and then keeps the remaining backbone frozen while updating only a small subset of parameters.

To address these issues, we present Hierarchical Exemplar Representation Adaptation (HERA), as shown in~\cref{fig:introduction}, an efficient VFM-based segmentation framework that learns from limited labels and adapts to novel domains without source data retraining. We first propose Hierarchical Layer Selection (HLS) to estimate a per-episode Exemplar Transfer Risk (ETR) for each candidate layer from the few labeled supports and the forward activations, and choose the lowest risk layer as the working representation. We then confine test time updates to a small subset of parameters at this layer and keep the rest of the backbone frozen. 

Even with a stable representation, token interaction maps, such as the self-attention maps used in ViTs, remain noisy under distribution shift. To mitigate this, Prior Guided Regularization (PGR) regularizes the attention on the selected representation with a lightweight spatial prior, strengthening locality and structural consistency while preserving global coverage. Finally, Pixelwise Adaptive Calibration (PAC) combines the selected representation with these prior guided attention maps and a query-image prototype-contrast map to calibrate pixel-level predictions, correcting residual artifacts along thin boundaries and in low-contrast regions. In summary, our contributions are:

\begin{itemize}
  \item We present HERA, a source-free test-time adaptation framework for CD-FSS with VFMs that organizes adaptation as select, regularize, and calibrate, keeps the backbone frozen while fine-tuning less than 2.7\% of parameters at test time in a lightweight manner.
\item We introduce Hierarchical Layer Selection (HLS), driven by a data\mbox{-}dependent Exemplar Transfer Risk (ETR), which selects the lowest\mbox{-}risk layer per episode from a few labeled supports and cached forward activations, thereby localizing adaptation to a small parameter subset.
\item We couple Prior\mbox{-}Guided Regularization (PGR) and Pixelwise Adaptive Calibration (PAC) to regularize target\mbox{-}side structure and calibrate pixel\mbox{-}wise predictions, yielding a coherent hierarchy from representation to prediction.
  \item Extensive experiments on multiple CD-FSS benchmarks show consistent gains over prior methods, improving by over 4.1 mIoU, with ablations validating each stage and confirming strong parameter efficiency.
\end{itemize}

%% file: sec/2_related.tex
\section{RELATED WORK}
\label{sec:RELATED WORK}

\textbf{Few-Shot Semantic Segmentation (FSS)} aims to predict pixel-level masks for novel classes using only a few labeled supports. Most existing methods fall into two families. Prototype-based approaches form class prototypes in feature space and classify queries by similarity~\citep{dong2018few,li2021adaptive,lang2022learning,lang2023base}. Affinity-based approaches compute dense correlations or attention between support and query features to propagate context~\citep{lu2021simpler,min2021hypercorrelation,fan2022self,peng2023hierarchical}. Subsequent works strengthen FSS through multi-scale context aggregation, prototype refinement, and mask-level regularization~\citep{tian2020prior,peng2023hierarchical,chen2024cross}, yet most assume matched training and test distributions, leaving robustness to domain shift largely unexplored.

\vspace*{0.05in}
\noindent\textbf{Cross-Domain Few-Shot Segmentation (CD-FSS)} aims to generalize to novel classes in an unseen target domain using a few labeled support exemplars. Prior work is largely CNN-based and follows two lines. Source-side training performs domain generalization on source data before deployment~\citep{lei2022cross,su2024domain,chen2024cross,fan2025adapting}, which can be effective but requires continued access to source data and repeated retraining. Target domain supervised adaptation either mines correspondences across images or fine-tunes adapter modules using a few labeled supports~\citep{wang2022adaptive,herzog2024adapt,tong2024lightweight,nie2024cross}. Despite progress, these pipelines are costly or source dependent, and convolutional inductive biases limit long range reasoning and robustness under distribution shift. In contrast, we adopt an effective VFM-based segmentation framework without source-data retraining.

\vspace*{0.05in}
\noindent\textbf{Vision Foundation Models (VFMs)} are largely ViT-based backbones pre-trained at scale with self-supervised, weakly-supervised, or semi-supervised objectives~\citep{oquab2023dinov2,chen2024internvl,simeoni2025dinov3,kirillov2023segment}. Representative models include CLIP~\citep{radford2021learning} for image-text alignment, MAE and EVA02~\citep{he2022masked,fang2023eva,fang2024eva} for masked-image modeling, SAM~\citep{kirillov2023segment} for promptable segmentation, and DINO~\citep{oquab2023dinov2,simeoni2025dinov3} for self-distillation with strong objectness cues. These models provide transferable hierarchical features and often yield competitive segmentation with a frozen encoder. Specifically in DINOv3~\citep{simeoni2025dinov3}, intermediate layers present coarse-to-fine semantics and attention that is locally coherent and globally aware, which suits support–query matching. Under distribution shift, however, layer utility varies across episodes and correspondence becomes unstable, so fixed layer choices or uniform fine-tuning are unreliable. We therefore adopt ViT-based VFMs with per-episode selection of a stable layer, followed by hierarchical regularization and calibration.

\noindent\textbf{Test-Time Adaptation (TTA)} adapts deployed models to target data using unlabeled test samples~\citep{wang2020tent,jia2024tinytta,liang2025comprehensive}. Common routes minimize entropy or consistency, update statistics, such as BN re-estimation, apply whitening or stylization, and perform contrastive or clustering-based alignment for segmentation~\citep{wang2022continual,gong2022note,kang2024membn}. Although deployment-friendly, they optimize surrogate losses on queries, require sizable trainable subsets or lengthy per-image updates, and are weakly coupled to the episodic nature of CD-FSS. In parallel, parameter-efficient fine-tuning (PEFT) updates a small fraction of weights via adapters, prompts, or low-rank modules~\citep{han2024parameter,hu2022lora,xing2024survey,chen2022vision,chen2022adaptformer}, but for dense prediction it often targets single-level proxies and lacks episode-aware alignment. Our approach unifies TTA and PEFT by updating a small parameter subset at test time on a frozen backbone, guided by stable representation selection.

\begin{figure*}[!t]
  \centering
  \includegraphics[page=1,width=\linewidth]{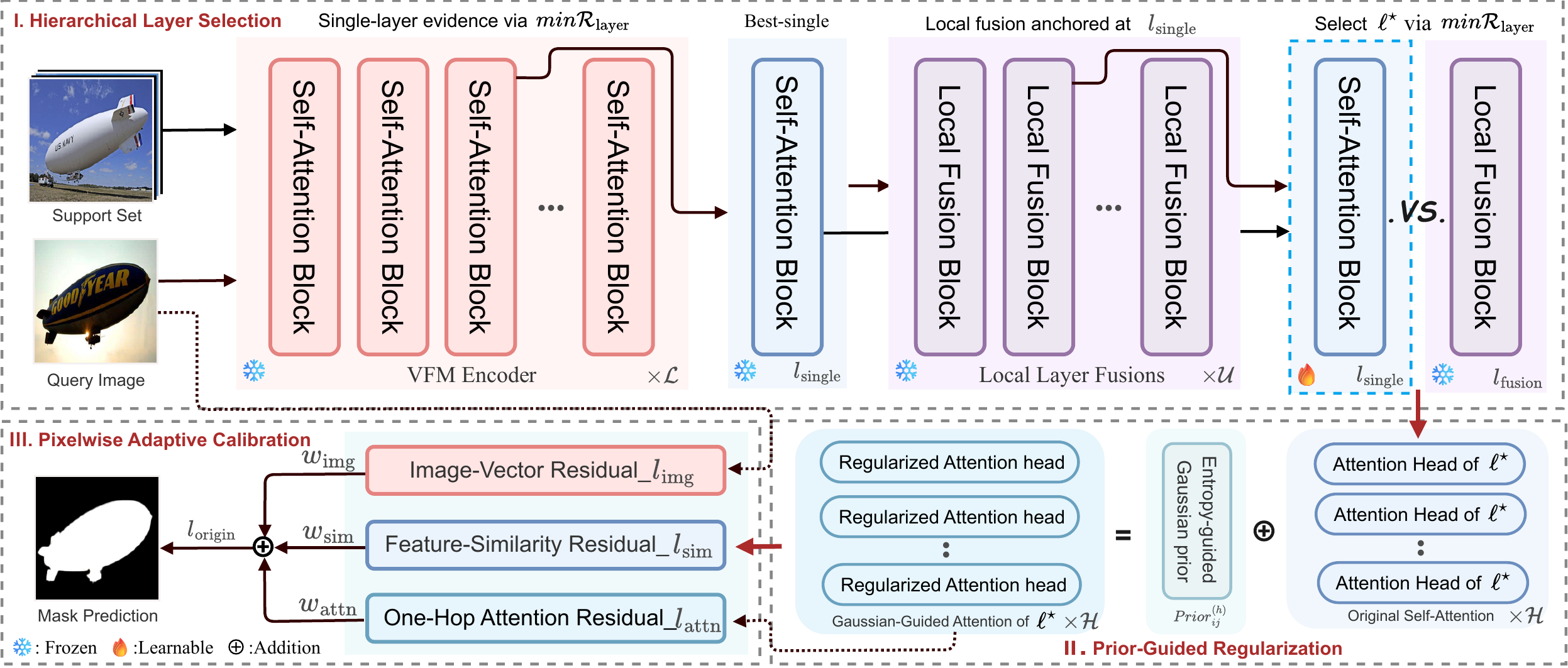}
  \vspace{-0.2in}
\caption{\small
HERA architecture. Hierarchical Layer Selection (HLS) estimates the leave-one-out layer risk $\mathcal{R}_{\mathrm{layer}}$ and routes each episode to $\ell^\star$, chosen from a single layer $\ell_{\mathrm{single}}$ or a local-fusion candidate $\ell_{\mathrm{fusion}}$. At $\ell^\star$, Prior Guided Regularization (PGR) calibrates self-attention with entropy-gated Gaussian priors, and Pixelwise Adaptive Calibration (PAC) adds residual logits $\ell_{\mathrm{sim}}, \ell_{\mathrm{attn}}, \ell_{\mathrm{img}}$ that are fused with the base logit in a select--regularize--calibrate hierarchy. $\mathcal{L}$, $\mathcal{U}$, and $\mathcal{H}$ are the numbers of layers, fusion candidates, and attention heads.}
  \label{fig:HERA-overview}
  \vspace*{-0.1in}
\end{figure*}

%% file: sec/3_method.tex
\section{Method}

\subsection{Architecture Overview}\label{architecture}
Cross-domain few-shot segmentation (CD-FSS) follows an episodic $K$-shot protocol: given a support--query set $\mathcal{S}=\{(I_s^i, M_s^i)\}_{i=1}^{K}$ and a query image $I_q$, models trained on source domains are evaluated on target domains with disjoint label spaces. Leveraging vision foundation models (VFMs) is attractive. However, under distribution shift, per-episode alignment between support and query becomes unstable, causing errors to cascade from representation to prediction. In addition, VFM transferability varies across layers, and local interactions remain noisy.

We therefore propose Hierarchical Exemplar Representation Adaptation (HERA), a three-stage select-regularize-calibrate framework that adapts at test time with a frozen backbone. We first design Hierarchical Layer Selection (HLS) to select a stable representation layer $\ell^{\ast}$ by minimizing a data-dependent Exemplar Transfer Risk (ETR) computed per episode. Then Prior Guided Regularization (PGR) refines the self-attention at $\ell^{\ast}$ with entropy-gated Gaussian priors, strengthening locality and structural consistency while preserving global coverage. Finally, Pixelwise Adaptive Calibration (PAC) calibrates pixel-wise predictions via lightweight residuals to recover thin structures and denoise low-contrast regions. Together, these stages form a hierarchical path from representation to prediction and transfer a few labeled supports into reliable guidance in new domains, yielding stable, well-calibrated predictions.

\subsection{Hierarchical Layer Selection}\label{sec:hlr}
Given a frozen ViT backbone $f$ that produces layerwise features $F^{\ell}$, we observe pronounced cross-layer transferability variance under distribution shift. We therefore make per-episode layer selection the primary stage of HERA. \text{HLS} minimizes the data-dependent episode-level ETR over candidate layers and outputs the selected representation $F^{\ell^\ast}$, which conditions downstream modules.

\subsubsection{Layerwise variability under shift}\label{sec:primary-risk}
\begin{figure*}[t]
  \centering
  \includegraphics[page=1,width=\linewidth]{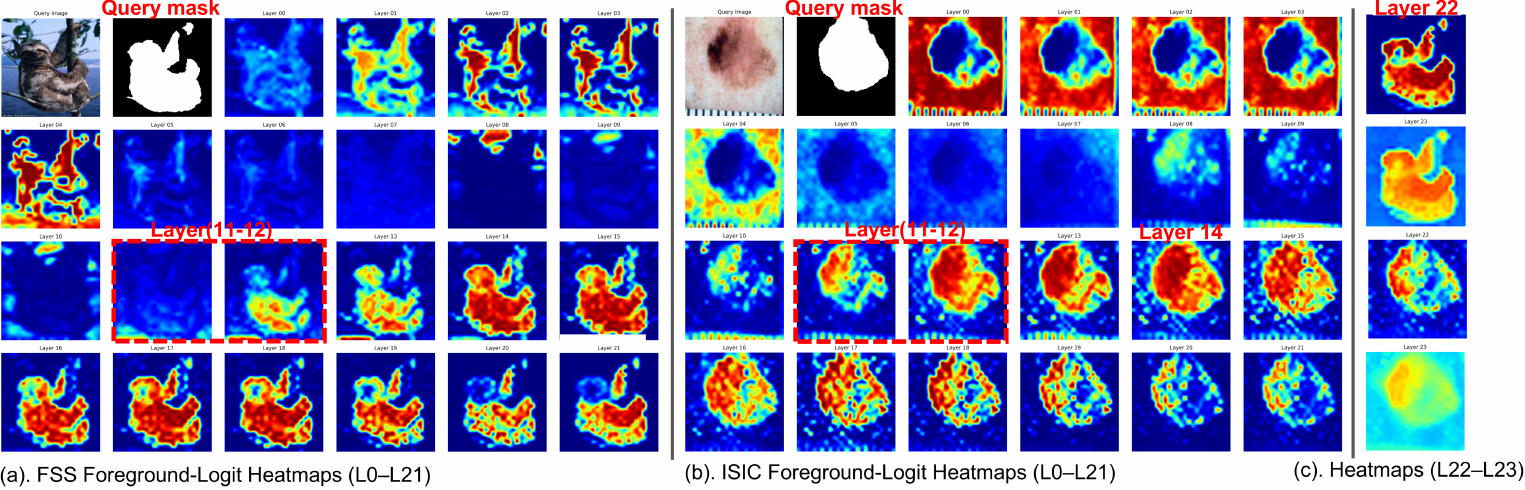}
  \vspace{-0.2in}
  \caption{\small \text{Layerwise variability in VFM features (DINOv3 example).}
  Foreground-logit heatmaps from ViT layers 0-23 for two episodes from distinct target domains. A semantic shift emerges around layers 11-12, yet the optimal layer for mask prediction differs across episodes, indicating episode- and domain-dependent variability in the best layer and motivating per-episode layer routing.}
  \label{fig:layer-variability}
\end{figure*}

As illustrated in \cref{fig:layer-variability}, early to mid layers 0-11 emphasize low-SNR edges and textures, whereas mid to high layers 12-23 yield class-agnostic objectness with sharper boundaries, with a consistent semantic shift around layers 11-12. Because the most informative layer varies across episodes and domains, any fixed choice is brittle. Episode-wise selection is thus crucial for robust support–query alignment. We therefore restrict routing to layers 12-23, where semantics consolidate while spatial detail is preserved. Single-layer routing in this band often performs well but is fragile on thin structures, occlusions, and clutter, as layers exhibit complementary failure modes. The optimal layer may fluctuate across episodes within a domain. To curb this instability, we augment the single-layer choice with a compact {local-fusion} candidate set centered at the best single layer and evaluate all candidates under a unified episode-level risk. This adds negligible cost and yields a more reliable routed representation for subsequent modules.

\subsubsection{Per-Episode Layer Selection}\label{sec:etr}
Following \cref{sec:primary-risk}, we estimate episode-level evidence at test time using leave-one-out and adapt only a small subset of parameters. Let the support set be $\mathcal{S}=\{(I_s^i,M_s^i)\}_{i=1}^{K}$. In the $i$-th iteration, $(I_s^i,M_s^i)$ acts as the pseudo-query $S_q^{i}$ and the remaining supports form $S^{(-i)}$. In the 1-shot setting, we synthesize two augmented views of the support for test time adaptation (TTA). Then we compute a prototype $\mathbf{P}_s^{\,i}$ from $S^{(-i)}$ and extract the pseudo-query feature $\mathbf{F}_q^{\,i}$ from $I_s^i$ at a candidate layer $\ell$.

We define Exemplar Transfer Risk (ETR) as one minus the average pseudo-query mIoU:

\begin{equation}\label{eq:sel-risk-1}
\mathcal{R}_{\text{layer}}(\ell)
  = 1 - \frac{1}{K}\sum_{i=1}^{K}
     \mathrm{mIoU}\!\Big(\cos(\mathbf{P}_s^{\,i},\,\mathbf{F}_q^{\,i}),\, M_q^{i}\Big),
\end{equation}
\begin{equation}\label{eq:sel-risk-2}
\ell^\star
  = \operatorname*{arg\,min}_{\ell \in \mathcal{C}} \ \mathcal{R}_{\text{layer}}(\ell),
\end{equation}
where $\mathcal{C}$ contains both single-layer and local-fusion candidates, and $\mathrm{mIoU}(\cdot)$ compares the predicted mask with the pseudo-query ground truth $M_q^{i}$, where $M_q^{i}$=$M_s^{i}$. The minimizer $\ell^\star$ is the routed layer used in subsequent stages.

With $\ell^\star$ fixed, we freeze the backbone and finetune only a small parameter set $\phi$ using the same leave-one-out construction, optimizing a binary segmentation loss:

\begin{equation}
\mathcal{L}_{\text{TTA}}
= \frac{1}{K}\sum_{i=1}^{K}
\mathrm{BCE}\!\Big(\!\cos(\mathbf{P}_{s}^{\,i,\ell^{\star}},\,\mathbf{F}_{q}^{\,i,\ell^{\star}}\big),\, M_q^{i}\Big),
\label{eq:ttf}
\end{equation}
where $\mathbf{P}_{s}^{\,i,\ell^\star}$ and $\mathbf{F}_{q}^{\,i,\ell^\star}$ are computed at the routed layer $\ell^\star$. To mitigate overfitting~\citep{he2020momentum,boudiaf2021few}, we fine-tune only the selected layer’s MLP.FC for single-layer routing and fusion-MLP.FC for local-fusion routing and all remaining parameters are frozen. See Appendices~7 and~10 for more details.

\begin{WideAlgo}{Hierarchical Layer Selection (HLS)}\label{alg:hls}
  \State \textbf{Require:} $K$-shot support set $\mathcal{S}$, candidate layers
  $\mathcal{L}=\{12,\ldots,23\}$ and \text{VFM} backbone $f$.
  \State Select trainable parameters
  \For{$n=1$ to $K{-}1$} \hfill $\lozenge$ \textit{\textbf{Episode-level routing evidence (n-shot): see \cref{sec:hlr}}}
    \State $\triangleright$ \textit{\textbf{Stage I: Single-layer evidence}}
    \State \quad Assemble the $n$-shot episode $S_n$; extract per-layer features $\{F^\ell\}_{\ell\in\mathcal{L}}$ with $f$
    \State \quad Compute $\mathcal{R}_{\text{layer}}(\ell)$ on $S_n$ and obtain $\ell_{\text{single}}$
           \hfill \textit{single risk, see \cref{eq:sel-risk-1}}
    \State $\triangleright$ \textit{\textbf{Stage II: Local fusion around $\ell_{\text{single}}$}}
    \State \quad Construct a local fusion pool $\mathcal{U}(\ell_{\text{single}})$
    \State \quad For each $U\in\mathcal{U}$, compute fusion weights $w_\ell$ and fused feature $F^{U}$
           \hfill \textit{see \cref{eq:two-stage-selection-1,eq:two-stage-selection-2}}
    \State \quad Choose $\ell^\star=\arg\min_{\{\ell_{\text{single}}\}\cup\mathcal{U}(\ell_{\text{single}})} \mathcal{R}_{\text{layer}}(\ell)$
           and fix $F^{\ell^\star}$ \hfill \textit{unified risk, see \cref{eq:sel-risk-2}}
    \For{$i=1$ to $K$} \hfill $\lozenge$ \textit{\textbf{Pseudo–query cross-evaluation}}
      \State Form $\langle S^{(-i)},\, (I_s^i,M_s^i)\rangle$ at fixed $F^{\star}$
      \State Compute $\mathcal{L}_{\text{TTA}}^{(i,n)}$ over all $n$-support combinations
             \hfill \textit{see \cref{eq:ttf}}
    \EndFor
    \State Update $\phi$ by back-propagating the average loss $\mathcal{L}_{\text{TTA}}^{(n)}$; keep $f$ frozen
           \hfill $\lozenge$ \textit{\textbf{Parameter-efficient TTA}}
  \EndFor
\end{WideAlgo}

\subsubsection{Two-Stage Selection with Lightweight TTA}
As shown in Algorithm 1, we first determine the best single layer $\ell_{\text{single}}$ via leave-one-out evidence and then form a compact set of local-fusion candidates $U \subseteq \mathcal{L} = \{12,\ldots,23\}$ anchored at $\ell_{\text{single}}$. Empirically, we observe that the last ViT block tends to provide more category-consistent and globally coherent features with strong objectness. We therefore include $\ell_{23}$ in each local-fusion candidate as a global context anchor to compensate for occlusion and fragmented shapes. All candidates are evaluated using the same episode-level risk ETR in \cref{eq:sel-risk-1}.

For any candidate $U$, let $r_\ell=\mathcal{R}_{\text{layer}}(\ell)$ denote the single-layer selection risk. Define the distance $d(\ell,\ell_{23})=|\ell-\ell_{23}|$ and compute fusion weights and the fused feature as:

\begin{equation}\label{eq:two-stage-selection-1}
w_\ell =
 \frac{\exp\!\big(-\beta r_\ell - \operatorname{dist}(\ell,\ell_{23})/\tau\big)}
      {\sum_{j\in U}\exp\!\big(-\beta r_j - \operatorname{dist}(j,\ell_{23})/\tau\big)},
\end{equation}
\begin{equation}\label{eq:two-stage-selection-2}
F^{\,U} = \sum_{\ell\in U} w_\ell\, F^{\,\ell},
\end{equation}
where $\beta>0$ controls reliance on the evidence $r_\ell$ and $\tau>0$ is a locality bandwidth that favors deeper, semantically aggregated layers. As $\tau\!\to\!\infty$, the locality term vanishes; as $\beta\!\to\!\infty$, the weights collapse to $\arg\min_{\ell\in U} r_\ell$, approaching the single-layer case. When evidence spreads across adjacent layers, a moderate $\tau$ trades off evidence and aggregation, reducing routing instability.

Fixing the routed layer at $\ell^\star$ mitigates the layer-wise component of episode-wise matching risk, thereby providing a more stable, episode-conditioned representation for downstream adaptation stages.

\subsection{Prior Guided Regularization}\label{sec:gga}
\begin{figure}[h]
  \centering
  \includegraphics[page=1,width=\linewidth]{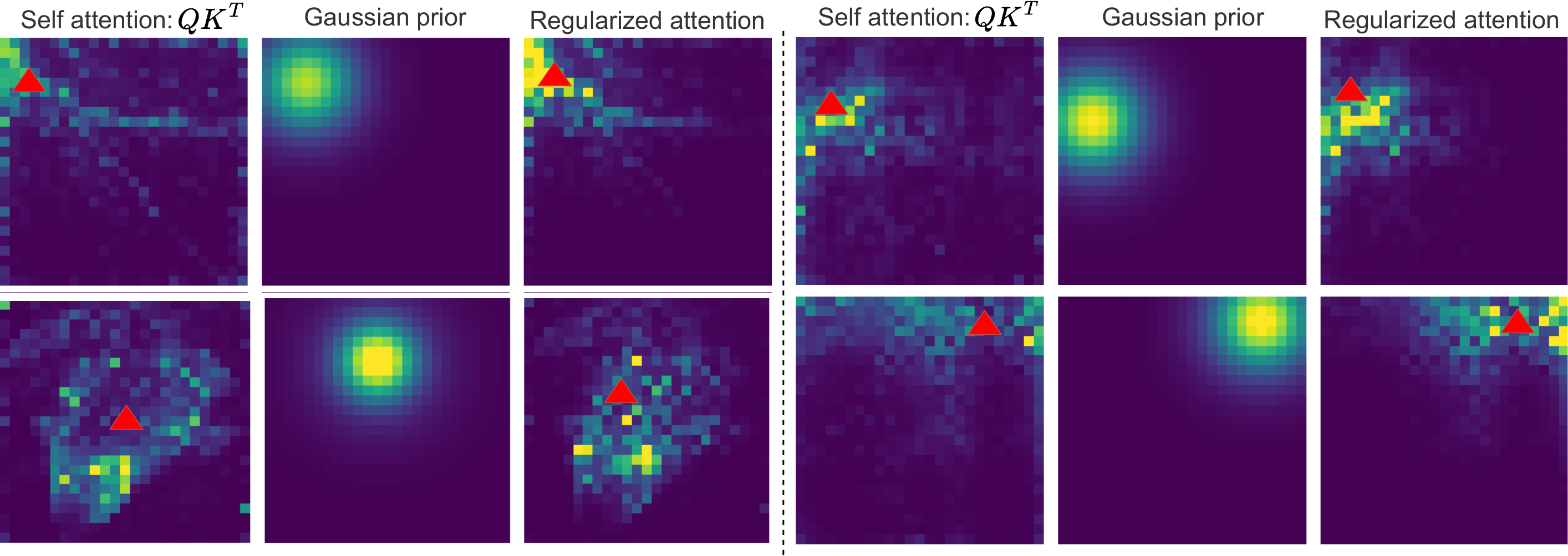}
  \vspace{-0.2in}
  \caption{\small \text{Prior Guided Regularization (PGR).}
  Per-head Gaussian priors, gated by entropy, are injected into $QK^{\top}$ logits to calibrate self-attention at $\ell^\star$,  locality is strengthened, far-field spurious peaks are suppressed, while preserving global coverage.}
  \label{fig:gga-episodes}
  \vspace{-0.2in}
\end{figure}

With the routed layer fixed by HLS, the representation provides stable global semantics, yet head-specific self-attention remains noisy under shift. Because DINO features provide mainly layer-level guidance, head-level maps show spurious long-range links, insufficient coverage of nearby neighborhoods, thin boundaries, and strong cross-head heterogeneity. A uniform, head-agnostic prior is therefore inadequate. We calibrate attention per head using a query-centered Gaussian prior whose bandwidth is set by an entropy gate derived from the head’s attention. Local and confident heads receive a sharper prior, whereas globally dispersed heads receive a more diffuse prior. This head-wise, entropy-gated calibration enforces locality while preserving necessary global coverage as shown in \cref{fig:gga-episodes}

\paragraph{Head-wise Gaussian Prior with Entropy Gating.}\vspace{-0.6em}
Attention heads in {ViT}s exhibit specialization in spatial scope and semantics ~\citep{raghu2021vision,lee2022surgical,roh2024levi}. We therefore impose a head-wise Gaussian prior and set its bandwidth by an entropy gate, yielding a {sharper prior} for local, confident heads and a {more diffuse prior} for globally dispersed ones:

\vspace{-0.1in}
\begin{equation}
\phi(p_j;\,p_i,\sigma)=\exp\!\Big(-\frac{\lVert p_j-p_i\rVert^{2}}{2\sigma^{2}}\Big),
\label{eq:gga-gauss}
\end{equation}
with two fixed bandwidth hyperparameters, $\sigma_{\mathrm{loc}} < \sigma_{\mathrm{glo}}$, capturing local aggregation and global coverage.

\begin{table*}[!t]
\centering
\footnotesize
\setlength{\tabcolsep}{2.6pt}
\caption{Quantitative comparison on the CD-FSS benchmark.
All compared methods, except HERA, are trained on the Pascal VOC~\citep{everingham2010pascal} source domain and then evaluated on four targets,
whereas HERA is source-free and performs no source- or target-side training (test-time adaptation only).
Best and second-best results are shown in \textbf{bold} and \underline{underlined}.
The \textsuperscript{\dag} indicates results reproduced by us, and \textsuperscript{\ddag} indicates a ViT-base backbone.
In the Training (S/T) column,
\stYesYes{} denotes source-side pretraining and supervised target retraining;
\stYesNo{} denotes source-side pretraining only (no target retraining);
\textbf{\stNone{}} denotes source-free, no target-side domain retraining, test-time adaptation only.}
\resizebox{\linewidth}{!}{%
\begin{tabular}{@{} M | P | T | cc cc cc cc || cc @{}}
\toprule
\multirow{2}{*}{Methods} & \multirow{2}{*}{Publication} & \multirow{2}{*}{Training (S/T)} &
\multicolumn{2}{c}{DeepGlobe} & \multicolumn{2}{c}{ISIC2018} &
\multicolumn{2}{c}{Chest X-ray} & \multicolumn{2}{c||}{FSS-1000} & \multicolumn{2}{c}{mIoU} \\
\cmidrule(lr){4-5}\cmidrule(lr){6-7}\cmidrule(lr){8-9}\cmidrule(lr){10-11}\cmidrule(lr){12-13}
& & & 1-shot & 5-shot & 1-shot & 5-shot & 1-shot & 5-shot & 1-shot & 5-shot & 1-shot & 5-shot \\
\midrule
PGNet~\citep{zhang2019pyramid}                   & ICCV 2019  & \stYesNo{} & 10.7 & 12.4 & 21.9 & 21.3 & 34.0 & 23.0 & 62.4 & 62.7 & 32.2 & 31.1 \\
PANet~\citep{wang2019panet}                   & ICCV 2019  & \stYesNo{} & 36.6 & 45.4 & 25.3 & 34.0 & 57.8 & 69.3 & 69.2 & 71.7 & 47.2 & 55.1 \\
CaNet~\citep{zhang2019canet}                    & CVPR 2019  & \stYesNo{} & 22.3 & 23.1 & 25.2 & 28.2 & 28.4 & 28.6 & 70.7 & 72.0 & 36.6 & 38.0 \\
RPMMs~\citep{yang2020prototype}                    & ECCV 2020  & \stYesNo{} & 13.0 & 13.5 & 18.0 & 20.0 & 30.1 & 30.8 & 65.1 & 67.1 & 31.6 & 32.9 \\
PFENet~\citep{tian2020prior}                   & TPAMI 2020 & \stYesNo{} & 16.9 & 18.0 & 23.5 & 23.8 & 27.2 & 27.6 & 70.9 & 70.5 & 34.6 & 35.0 \\
RePRI~\citep{boudiaf2021few}                    & CVPR 2021  & \stYesNo{} & 25.0 & 27.4 & 23.3 & 26.2 & 65.1 & 65.5 & 71.0 & 74.2 & 46.1 & 48.3 \\
HSNet~\citep{min2021hypercorrelation}                    & ICCV 2021  & \stYesNo{} & 29.7 & 35.1 & 31.2 & 35.1 & 51.9 & 54.4 & 77.5 & 81.0 & 47.6 & 51.4 \\
SSP\textsuperscript{\dag}~\citep{fan2022self} & ECCV 2022 & \stYesNo{} & 40.5 & 49.6 & 35.5 & 48.2 & 74.2 & 74.5 & 79.0 & 80.2 & 57.3 & 63.1 \\
DATO~\citep{li2025dual}                       & CVPR 2025 & \stYesNo{} & \best{51.1} & \secbest{59.3} & \best{68.8} & \secbest{70.3} & 79.6 & 81.1 & \secbest{81.8} & 84.6 & \best{70.3} & \secbest{73.8} \\
\midrule
PATNet~\citep{lei2022cross}                   & ECCV 2022  & \stYesYes{} & 37.9 & 43.0 & 41.2 & 53.6 & 66.6 & 70.2 & 78.6 & 81.2 & 56.1 & 62.0 \\
PMNet~\citep{chen2024pixel}                    & WACV 2024  & \stYesYes{} & 37.1 & 41.6 & 51.2 & 54.5 & 70.4 & 74.0 & \best{84.6} & \secbest{86.3} & 60.8 & 64.1 \\
ABCDFSS~\citep{herzog2024adapt}                  & CVPR 2024  & \stYesYes{} & 42.6 & 49.0 & 45.7 & 53.3 & 79.8 & 81.4 & 74.6 & 76.2 & 60.7 & 65.0 \\
APSeg\textsuperscript{\ddag}~\citep{he2024apseg}  & CVPR 2024 & \stYesYes{} & 35.9 & 40.0 & 45.4 & 54.0 & 84.1 & 84.5 & 79.7 & 81.9 & 61.3 & 65.1 \\
DR-Adapter~\citep{su2024domain}               & CVPR 2024  & \stYesYes{} & 41.3 & 50.1 & 40.8 & 48.9 & 82.4 & 82.3 & 79.1 & 80.4 & 60.9 & 65.4 \\
APM ~\citep{tong2024lightweight}                      & NeurIPS 2024 & \stYesYes{} & 40.9 & 44.9 & 41.7 & 51.2 & 78.3 & 82.8 & 79.3 & 81.9 & 60.0 & 65.2 \\
IFA~\citep{nie2024cross}                      & CVPR 2024  & \stYesYes{} & \secbest{50.6} & 58.8 & \secbest{66.3} & 69.8 & 74.0 & 74.6 & 80.1 & 82.4 & 67.8 & 71.4 \\
TVGTANet~\citep{liu2025textual}               & ACM MM2025 & \stYesYes{} & 42.0 & 50.7 & 47.2 & 58.8 & \secbest{84.6} & \secbest{87.3} & 78.3 & 81.4 & 63.0 & 69.5 \\
SDRC\textsuperscript{\ddag}~\citep{tong2025self}                     & ICML 2025 & \stYesYes{} & 43.2 & 46.8 & 46.6 & 55.0 & 82.9 & 84.8 & 80.3 & 82.6 & 63.2 & 67.3 \\
DFN~\citep{tong2025adapter}                   & ICML 2025  & \stYesYes{} & 39.5 & 47.7 & 50.4 & 58.5 & 83.2 & 87.1 & 83.0 & 85.7 & 64.0 & 69.8 \\
LoEC\textsuperscript{\ddag}~\citep{liu2025devil}  & CVPR 2025  & \stYesYes{} & 42.1 & 51.5 & 52.9 & 62.4 & 83.9 & 84.1 & 81.1 & 83.7 & 65.0 & 70.4 \\
\midrule
HERA\textsuperscript{\ddag} (DINOv2)
                        & ----         & \textbf{\stNone{}} & 41.2 & 57.8 & 55.6 & 68.7 & 83.2 & 86.9 & 70.2 & 80.3 & 62.6 & 73.4 \\
HERA\textsuperscript{\ddag} {(DINOv3)}
                        & ----         & \textbf{\stNone{}} & 44.6 & \best{63.4} & 61.2 & \best{73.6} & \best{85.8} & \best{87.9} & 81.6 & \best{86.7} & \secbest{68.3} & \best{77.9} \\
\bottomrule
\end{tabular}}
\label{tab:main}
\end{table*}

Let $\bar H_q^{(h)}$ denote the mean row entropy of the $QK^{\top}$ logits at head $h$, indicating global dispersion, and $\bar H_k^{(h)}$ that of $KK^{\top}$, indicating local stability. Using a logistic gate $g(\cdot)$ with temperature $\alpha>0$:
\begin{equation}
\begin{aligned}
\gamma_h &= g\!\big(\alpha(\bar H_q^{(h)}-\bar H_k^{(h)})\big),\\
\sigma_h &= (1-\gamma_h)\,\sigma_{\mathrm{glo}}+\gamma_h\,\sigma_{\mathrm{loc}},
\end{aligned}
\label{eq:gga-gate}
\end{equation}
where heads with stronger locality, indicating larger $\bar H_q^{(h)}-\bar H_k^{(h)}$, receive a sharper prior, and globally dispersed heads receive a more diffuse prior.

\subsection{Pixelwise Adaptive Calibration}
With the layer and attention stages in place, pixelwise decisions remain unstable under shift, leaving residual errors along thin boundaries and in low-contrast regions. PAC explicitly performs pixelwise calibration by fusing three lightweight cues computed from $F^{\ell^\ast}$ and the refined attention, namely feature similarity, one-hop attention propagation, and image appearance. This fusion corrects residual artifacts and yields consistent masks:
\begin{multline}
\ell_{\mathrm{final}}(x)=\ell_{0}(x)+w_{\mathrm{sim}}\ell_{\mathrm{sim}}(x)\\[-2pt]
+\,w_{\mathrm{attn}}\ell_{\mathrm{attn}}(x)+w_{\mathrm{img}}\ell_{\mathrm{img}}(x)
\label{eq:pac}
\end{multline}
where $\ell_{0}(x)$ is the base logit from the selected representation and $w_{\cdot}$ are fixed scalar weights. A single-step refine-vote gate applies residuals only when the estimated gain is positive, adding negligible overhead. Taken together, the three stages realize a hierarchical select-regularize-calibrate pipeline that adapts at test time with a frozen backbone. See Appendix~5 for more details.

%% file: sec/4_experiments.tex
\section{Experiments}

\vspace*{-0.05in}
\noindent{\bf Evaluation Datasets and Metrics.} We evaluate in a source-free test-time adaptation setting without access to source data. We evaluate on four public target datasets: FSS-1000~\citep{li2020fss}, a natural image corpus spanning 1{,}000 object classes; DeepGlobe~\citep{demir2018deepglobe}, a satellite land cover dataset with seven categories and pronounced texture and scale shifts; ISIC2018~\citep{tschandl2018ham10000,codella2019skin}, a dermoscopic skin lesion dataset with low-contrast and irregular boundaries; and Chest X-ray~\citep{jaeger2013automatic,candemir2013lung}, a medical radiograph dataset with substantial grayscale and structural variation. We follow the standard episodic K-shot protocol and report the mean IoU for 1-shot and 5-shot regimes. To ensure comparability, all methods use identical episode sampling, a shared preprocessing pipeline, and a unified input resolution of $400\times400$.

\begin{figure*}[t]
  \centering
  \includegraphics[page=1,width=\linewidth]{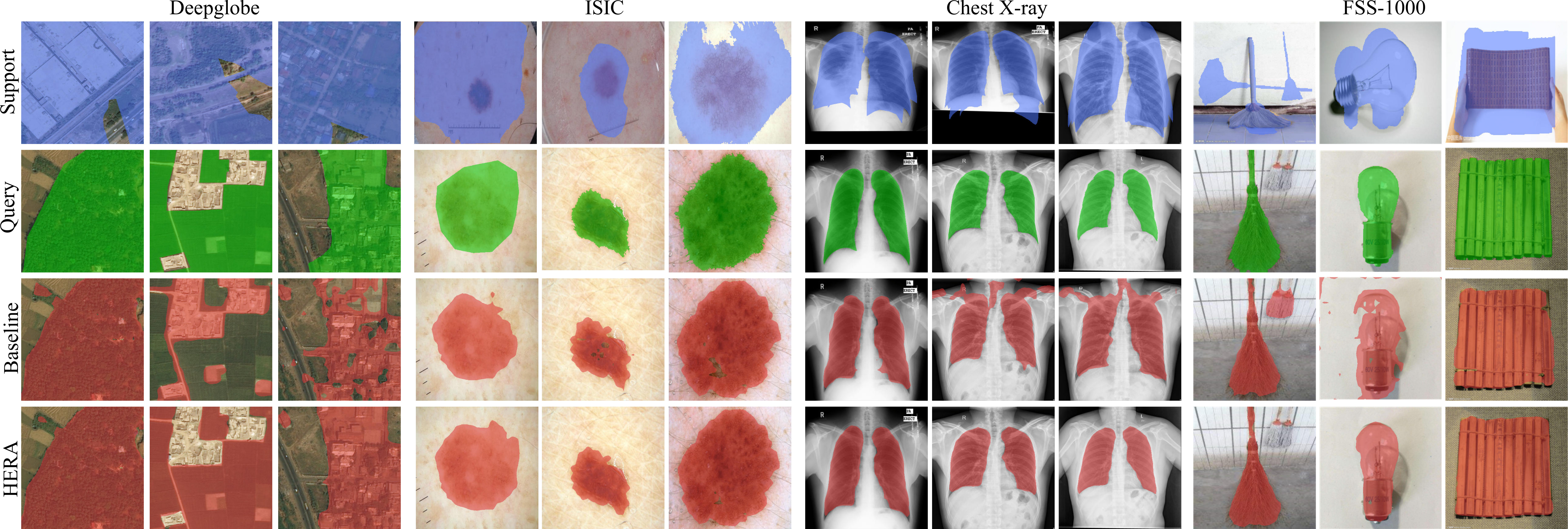}
  \caption{\small Qualitative results on the Chest X-ray, ISIC, FSS-1000,
and Deepglobe datasets under the 1-shot setting. The prediction and ground truth of query images are in red and green, respectively. Best viewed in color.}
  \label{fig:Qualitative}
  \vspace{-0.5em}
\end{figure*}

\noindent\textbf{Implementation Details.}
We use SSP~\citep{fan2022self} as the few-shot head and run HERA on a ViT backbone (default DINOv3~\citep{simeoni2025dinov3}).
Under test-time adaptation (TTA), each target episode proceeds as follows:
(i) HLS selects the routed layer; and
(ii) we form leave-one-out splits of the $K$ supports and minimize the loss in~\eqref{eq:ttf} on these splits, performing $K{-}1$ lightweight updates.
We optimize with Adam (learning rate $1.3\times10^{-3}$, $\beta_1=0.9$, $\beta_2=0.999$).
Only a small parameter subset is trainable: the single-layer variant updates that layer’s {mlp.fc}, and the local-fusion variant updates the {fusion-mlp.fc} and all other weights remain frozen.
In the 1-shot setting, we synthesize {two} augmented views from the single support using a soft copy–paste scheme to stabilize TTA.
All experiments run on a single NVIDIA A100, with peak VRAM of \(4.2\) GB (\( 5\%\) of capacity). Since fewer than \(2.7\%\) of parameters are updated, both compute and memory overheads are negligible.

\subsection{Comparison Experiments}
Our HERA unifies VFM-based adaptation for CD-FSS into a source-free, no-retraining test-time pipeline that is fundamentally different from the training paradigms of the methods reported in Table~\ref{tab:main}. Under an identical evaluation protocol with the same episode sampling, preprocessing, and \(400{\times}400\) input, HERA with DINOv3 attains 68.3/77.9 mIoU in the 1-shot/5-shot settings. It outperforms LoEC~\citep{liu2025devil} by +3.3/+7.5, SDRC~\citep{tong2025self} by +5.1/+10.6, and the no-retraining baseline SSP by +11.0/+14.8 mIoU. Even compared with the CNN-based DATO~\citep{li2025dual}, which requires source-domain training, HERA leads by +4.1 in the 5-shot setting and exceeds the 5-shot performance of all other CNN-based methods. At the same time, HERA remains source-free, which better satisfies privacy and engineering constraints and simplifies deployment.

In the 1-shot setting, HERA also provides practical advantages in both performance and efficiency as shown in~\cref{fig:Qualitative}. It surpasses TVGTANet~\citep{liu2025textual} and DFN~\citep{tong2025adapter} by +5.3 and +4.3 mIoU, respectively, even though these methods require training on both source and target domains. The proposed denoising and boundary calibration modules are specifically designed for HERA and yield natural, well-defined boundaries. On Chest X-ray, HERA achieves the best results with 85.8/87.9 mIoU, indicating the combination of selection–regularization–calibration is particularly effective for low-contrast, thin-boundary clinical cases. 

\begin{table}[t]
\centering
\huge
\caption{HERA component ablation (5-shot mIoU$\uparrow$, frozen VFM, $400{\times}400$). 
HLS/PGR/PAC as defined; ‘$\Delta$ Avg.’ = gain over SSP.}
\label{tab:ablate-main-indep}
\setlength{\tabcolsep}{6pt}
\resizebox{\columnwidth}{!}{
{\renewcommand{\arraystretch}{1.2}%
\begin{tabular}{l|cccc|cc}
\toprule
Method & DeepGlobe & ISIC & Chest X-ray & FSS-1000 & 5-shot & $\Delta$ Avg. \\
\midrule
Baseline            & 49.6 & 48.2 & 74.5 & 80.2 & 63.1 & \phantom{+}0.0 \\
+ HLS               & 61.7 & 71.4 & 87.7 & 86.0 & 76.7 & +13.6 \\
+ HLS + PGR         & \underline{62.6} & \underline{72.0} & \underline{88.0} & 86.5 & \underline{77.3} & \underline{+14.2} \\
+ HLS + PAC         & 62.1 & 71.6 & \textbf{88.3} & \underline{86.6} & 77.2 & +14.1 \\
\textbf{+ HLS + PGR + PAC}
                    & \textbf{63.4} & \textbf{73.6} & 87.9 & \textbf{86.7} & \textbf{77.9} & \textbf{+14.8} \\
\bottomrule
\end{tabular}}}
\end{table}

\begin{table}[t]
  \centering
  \small
  \setlength{\tabcolsep}{4pt}
  \caption{PAC branch ablation on HLS+PGR (5-shot mIoU$\uparrow$, frozen VFM). 
 ‘$\Delta$ Avg.’ = gain vs. HLS+PGR.}
  \label{tab:ablate-refine-indep}
  {\renewcommand{\arraystretch}{1.15}%
  \begin{tabular*}{0.95\columnwidth}{@{\extracolsep{\fill}}lcc}
    \toprule
    Variant & 5-shot & $\Delta$ Avg. \\
    \midrule
    Baseline + HLS + PGR                         & 77.27 & --    \\
    + l\text{-}sim                               & \underline{77.57} & \underline{+0.30} \\
    + l\text{-}attn                              & 77.49 & +0.22 \\
    + l\text{-}img                               & 77.45 & +0.18 \\
    \textbf{+ l\text{-}sim + l\text{-}attn + l\text{-}img}
                                                 & \textbf{77.91} & \textbf{+0.64} \\
    \bottomrule
  \end{tabular*}
  }
\end{table}

Overall, HERA delivers state-of-the-art performance in the 5-shot setting without any source or target training and remains strong in the 1-shot setting, while updating only \textbf{$<2.7\%$} of its parameters. This yields a favorable trade-off between accuracy and cost, and makes deployment across institutions and devices easier and less constrained. Analyses with other VFMs are provided in Appendix~8.

\begin{table*}[t]
\centering
\footnotesize
\setlength{\tabcolsep}{4pt}
\caption{Layer selection ablation with a frozen backbone (5-shot mIoU$\uparrow$). 
Rule lists the per episode selector and notation includes $\mathbf{g}_\ell$ gradient w.r.t.\ features of layer $\ell$; $S_{\text{sem}}$, $S_{\text{str}}$, $C$ normalized semantic, structure, and complexity scores; and $\mathrm{mIoU}_{\text{sup}}(\ell)$ support only mIoU.}
\label{tab:ablate_hlr_selection}
{\renewcommand{\arraystretch}{1.3}%
\begin{tabular*}{0.95\textwidth}{@{\extracolsep{\fill}}cc|C{1.6cm}C{1.6cm}C{1.6cm}C{1.6cm}|cc}
\toprule
Method & Rule & DeepGlobe & ISIC & Chest X\text{-}ray & FSS\text{-}1000 & Avg. & $\Delta$ \\
\midrule
Static-Max
& $\arg\max_{\ell}\big(\alpha\,S_{\text{sem}}(\ell)+\beta\,S_{\text{str}}(\ell)+\gamma\,C(\ell)\big)$
& 58.8 & \underline{67.2} & 80.1 & 81.4 & 71.9 & \phantom{+}0.0 \\
Grad\text{-}Max 
& $\arg\max_{\ell}\ \lVert \mathbf{g}_{\ell}\rVert$
& 60.2 & 62.0 & 85.4 & 84.7 & 73.1 & +1.2 \\
Grad$\Delta$\text{-}Max 
& $\arg\max_{\ell}\ \big\lVert \mathbf{g}_{\ell}-\mathbf{g}_{\ell-1}\big\rVert$
& \underline{60.4} & 61.7 & \underline{86.0} & \underline{84.8} & \underline{73.2} & \underline{+1.3} \\
\textbf{HLS (ETR)} 
& $\arg\min_{\ell}\big(1-\mathrm{mIoU}_{\text{sup}}(\ell)\big)$
& \textbf{61.7} & \textbf{71.4} & \textbf{87.7} & \textbf{86.0} & \textbf{76.7}  & \textbf{+4.8} \\
\bottomrule
\end{tabular*}
}
\end{table*}

\subsection{Ablation Studies}
\noindent\textbf{Component ablation.}
We ablate HERA in the 5-shot setting with a frozen VFM backbone as shown in Table~\ref{tab:ablate-main-indep}. The SSP baseline averages 63.1 mIoU, adding HLS raises the mean to 76.7 mIoU, which provides the dominant gain. With HLS fixed, attaching PGR or PAC raises the average to 77.3 and 77.2 mIoU, respectively. These improvements are positive and stable across all four datasets, with per-dataset gains of +0.2 to +0.9 mIoU on average. The full-stack HLS+PGR+PAC reaches 77.9 mIoU, a gain of total +14.8 mIoU over SSP. Qualitatively, PGR suppresses spurious far-field correlations and strengthens local coherence, whereas PAC calibrates pixels in low-contrast, thin-boundary regions, yielding cleaner borders and less leakage. These effects also reduce failures in the 1-shot regime, and the marginal cost of adding these modules is minimal. We update $<2.7\%$ of the parameters, and the additional computational overhead is negligible. Overall, HLS aligns the routed representation, while PGR and PAC provide robust, interpretable structural refinements, yielding a cumulative top-down select-regularize-calibrate effect.

\noindent\textbf{Interaction of PGR and PAC Components.}
With HLS at 76.7 mIoU, adding PGR raises the mean to 77.3 mIoU, adding PAC yields 77.2 mIoU, and enabling both reaches 77.9 mIoU. The corresponding gains over HLS are 0.6, 0.5, and 1.2 mIoU, see Table~\ref{tab:ablate-main-indep}, which is slightly above 1.1 and indicates complementarity rather than redundancy. Decomposing PAC in Table~\ref{tab:ablate-refine-indep} relative to HLS+PGR, the similarity residual $\ell_{\mathrm{sim}}$, the one-hop attention prior $\ell_{\mathrm{attn}}$, and the image cue $\ell_{\mathrm{img}}$ add 0.30, 0.22, and 0.18 mIoU, respectively, and using all three reaches 77.91 mIoU, an additional 0.64 mIoU over HLS+PGR. PGR regularizes attention at the representation level, while PAC calibrates predictions at the pixel level, thereby making the two modules complementary and consistently effective.

\noindent\textbf{Effect of Per-Episode Layer Selection (HLS).}
We compare per-episode selectors for test-time adaptation with a frozen backbone, as shown in Table~\ref{tab:ablate_hlr_selection}. Our criterion-based HLS routes each episode to the layer that maximizes support-only mIoU, $\arg\min_{\ell}\bigl(1-\mathrm{mIoU}_{\mathrm{sup}}(\ell)\bigr)$, and provides task-aligned, episode-aware feedback without extra parameters or surrogate losses. In contrast, Static-Max scores feature quality using semantic, structural, and complexity heuristics rather than task fit. Gradient proxies, including Grad-Max and Grad$\Delta$-Max, rank layers by gradient magnitude or change. These proxies tend to favor the final blocks in ViTs because of residual connections and normalization, and they correlate weakly with semantic alignment. HLS attains the best average performance of 76.7 mIoU, surpassing Static-Max by +4.8 mIoU and the strongest gradient proxy by +3.5 mIoU. These results justify per-episode layer selection as the primary performance driver and the natural entry point of HERA. The gap is larger on other VFM backbones, for example, DINOv2~\citep{oquab2023dinov2}.


\noindent\textbf{Adaptation Cost in New Target Domains.}
Many retraining-based and domain-generalization methods require tens to hundreds of GPU-hours of computation on source data prior to deployment. In contrast, HERA follows a source-free, no-retraining test-time adaptation paradigm. For a new target domain, we run a single adaptation phase driven by the available supports and then reuse the adapted model for all remaining queries. Using a DINOv3 backbone in the 1-shot setting, the per-episode costs are 0.202 s for hierarchical layer selection, 0.280 s for the lightweight parameter update, and 0.243 s for inference, with only 8.39M trainable parameters (2.69\%). Each stage, therefore, adds only a small fraction of latency and memory relative to standard inference. This design allows HERA to quickly specialize a frozen VFM to a new domain while still delivering strong cross-domain performance.

%% file: sec/5_conclusion.tex
\section{Conclusion}
We identify layer-wise transferability variation and noisy head-level interactions under domain shift, rather than limited representational capacity, as the key bottlenecks when applying VFMs to CD-FSS. We introduce Hierarchical Exemplar Representation Adaptation (HERA), a three-stage select-regularize-calibrate VFM-based segmentation framework that learns effectively from limited labels and adapts to novel domains without source-data retraining. Across domains and backbones, HERA improves mIoU by over 4.1 while updating 2.69\% of its parameters, providing a low-overhead and practical way to deploy VFMs in CD-FSS.



%% file: sec/X_suppl.tex
\clearpage
\setcounter{page}{1}
\maketitlesupplementary

\section{CD-FSS Benchmark Datasets}
We conduct experiments on four target-domain datasets: FSS-1000~\citep{li2020fss}, DeepGlobe~\citep{demir2018deepglobe}, ISIC2018~\citep{tschandl2018ham10000,codella2019skin}, and Chest X-ray~\citep{candemir2013lung,jaeger2013automatic}.  
Together, these datasets cover natural images, satellite imagery, dermoscopic images, and radiological scans, and thus exhibit substantial variation in appearance, scale, and imaging modality.

\vspace{4pt}\noindent\textbf{FSS-1000}~\citep{li2020fss} is a few-shot semantic segmentation dataset containing 1000 object categories, each associated with 10 pixel-wise annotated images.  
Following the standard CD-FSS protocol, we adopt the official semantic segmentation split.  
We report results on the designated test subset, which includes 240 categories and 2400 images and serves as one of our target domains.

\vspace{4pt}\noindent\textbf{DeepGlobe}~\citep{demir2018deepglobe} is a satellite land-cover dataset with dense per-pixel annotations for seven semantic classes: urban, agriculture, rangeland, forest, water, barren, and unknown.  
Because labels are only available for the training partition, we use the official training set, comprising 803 annotated images, as the target-domain data for evaluation.

\vspace{4pt}\noindent\textbf{ISIC2018}~\citep{tschandl2018ham10000,codella2019skin} is a dermoscopic skin-lesion segmentation benchmark designed for melanoma screening.  
The dataset contains high-resolution dermoscopy images, each with an expert-annotated binary mask for the primary lesion region.  
We follow the preprocessing and data splits commonly adopted in previous work and treat ISIC2018 as a medical target domain characterised by low contrast and irregular object boundaries.

\vspace{4pt}\noindent\textbf{Chest X-ray}~\citep{candemir2013lung,jaeger2013automatic} is a tuberculosis screening dataset composed of 566 high-resolution posterior–anterior chest radiographs (approximately $4020 \times 4892$ pixels).  
The images were collected from 58 tuberculosis cases and 80 normal cases.  
To make training feasible and ensure consistency, we resize each radiograph to $1024 \times 1024$ pixels and use this dataset as a grayscale medical target domain.

\section{Details of SSP Baseline}
The SSP~\citep{fan2022self} baseline is a prototype-based few-shot semantic segmentation approach that refines the support prototype by exploiting the query image. 
Let $\hat{F}_q \in \mathbb{R}^{c\times h\times w}$ denote the enhanced query feature map and $P_s$ denote the foreground support prototype.
SSP first produces coarse foreground and background probability maps by matching $\hat{F}_q$ to $P_s$ using cosine similarity:
\begin{equation}
M_{\mathrm{coarse}} = \mathrm{softmax}\big(\cos(\hat{F}_q, P_s)\big),
\end{equation}
where $\cos(\cdot,\cdot)$ computes channel-wise cosine similarity between $P_s$ and each spatial feature of $\hat{F}q$, and the softmax is applied over the foreground/background dimension, yielding $M{\mathrm{coarse}}={M_{\mathrm{coarse}}^{b}, M_{\mathrm{coarse}}^{f}}\in\mathbb{R}^{2\times h\times w}$.

\vspace{4pt}\noindent To obtain a query-adapted foreground prototype, SSP aggregates high-confidence foreground locations from $\hat{F}_q$ via masked average pooling:
\begin{equation}
P_q^{f} = \mathrm{MAP}\big(\hat{F}_q, M_{\mathrm{coarse}}^{f}(x,y) > \tau_f\big),
\end{equation}
where $\mathrm{MAP}(\cdot)$ denotes masked average pooling over spatial positions $(x,y)$ whose foreground probability exceeds $\tau_f = 0.7$ and the resulting prototype satisfies $P_q^{f}\in\mathbb{R}^{c}$.

\vspace{4pt}\noindent To model complex background regions, SSP constructs adaptive self-support background prototypes.  
We first compute a masked background feature map as:
\begin{equation}
\hat{F}_q^{b} = \hat{F}_q \odot \mathbb{I}\big(M_{\mathrm{coarse}}^{b}(x,y) > \tau_b\big),
\end{equation}
where $\odot$ denotes element-wise multiplication, $\mathbb{I}(\cdot)$ is the indicator function, and $\tau_b = 0.6$ is the confidence threshold for background pixels.  
As a result, $\hat{F}_q^{b}$ only keeps features at reliable background positions.

\vspace{4pt}\noindent Next, we reshape $\hat{F}_q^{b}$ and $\hat{F}_q$ into matrices in $\mathbb{R}^{c\times t}$ and $\mathbb{R}^{c\times n}$, respectively, where $t$ is the number of selected background locations and $n=h\times w$ is the total number of spatial positions.  
The affinity between background features and all query features is computed as:
\begin{equation}
A = \mathrm{matmul}\big(\hat{F}_q^{b\top}, \hat{F}_q\big),
\end{equation}
where $\mathrm{matmul}(\cdot,\cdot)$ denotes matrix multiplication, producing a similarity matrix $A\in\mathbb{R}^{t\times n}$ whose $(i,j)$\textsuperscript{th} entry measures the similarity between the $i$th background feature and the $j$th query location.

\vspace{4pt}\noindent Adaptive background prototypes are derived by attention-style aggregation:
\begin{equation}
P_q^{b} = \mathrm{matmul}\!\big(\hat{F}_q^{b}, \mathrm{softmax}(A)\big),
\end{equation}
where the softmax is applied along the first dimension of $A$ so that each query location attends to a normalised weight distribution over the $t$ background features.  
The matrix multiplication then computes the corresponding weighted sums, which yield query-dependent background prototypes $P_q^{b}\in\mathbb{R}^{c\times n}$. And then they are further pooled into a compact representation.

\vspace{4pt}\noindent Finally, SSP combines the original support prototype with the self-support prototypes to form the fused prototype set:
\begin{equation}
P = \alpha_1 P_s + \alpha_2 P_q, \qquad P_q = \{P_q^{f}, P_q^{b}\},
\end{equation}
where $\alpha_1=\alpha_2=0.5$ are fixed fusion weights, $P_s$ is the support-derived prototype, and $P_q$ collects the query-derived foreground and background prototypes.

\vspace{4pt}\noindent The refined segmentation for the query image is then obtained by matching $\hat{F}_q$ with the fused prototype set:
\begin{equation}
\bar{M}_q = \mathrm{softmax}\big(\cos(\hat{F}_q, P)\big),
\end{equation}
where $\bar{M}_q\in\mathbb{R}^{2\times h\times w}$ denotes the final foreground/background probability maps obtained by applying a channel-wise softmax to the cosine-similarity scores.

\section{Ablation on MLP Components}\label{app:mlp_ablation}
\paragraph{MLP head variants on the selected layer.}
With the backbone frozen and the working representation layer chosen by HLS, we attach a small MLP head at that layer to improve support-query correspondence at test time.
We evaluate three variants: {M0}, no MLP head ({apply\_fc}=False); {M1}, MLP branch present but frozen ({apply\_fc}=True, {zero\_init}=True; parameters fixed); and {M2}, a trainable MLP fine-tuned at test time on the selected layer ({apply\_fc}=True).
Only the MLP head is updated, keeping the fraction of updated parameters below 2.7\%.

\begin{table}[h]
  \centering
  \captionsetup{width=\linewidth}
  \caption{MLP ablation at the layer selected by HLS with the backbone frozen. $\Delta$ denotes the improvement relative to the row above.}
  \label{tab:mlp_ablation}
  \begin{minipage}{\linewidth}
    \centering
    \small
    \setlength{\tabcolsep}{6pt}
    \begin{tabular*}{\linewidth}{@{\extracolsep{\fill}}lcccc@{}}
      \toprule
      Variant & mIoU@1 & $\Delta$ & mIoU@5 & $\Delta$ \\
      \midrule
      M0: no MLP        & 65.66 & --    & 75.20 & --    \\
      M1: MLP frozen    & 66.33 & +0.67 & 75.78 & +0.58 \\
      M2: trainable MLP & \bfseries 68.29 & \bfseries +1.96 &
                          \bfseries 77.91 & \bfseries +2.13 \\
      \bottomrule
    \end{tabular*}
  \end{minipage}
\end{table}

\vspace{-10pt}\paragraph{Analysis of MLP ablation.}
Starting from M0 at 65.66 mIoU in one shot and 75.20 mIoU in 5-shot, as shown in Table~\ref{tab:mlp_ablation}, adding a frozen residual MLP branch (M1) raises the means to 66.33 and 75.78 mIoU, with gains of 0.67 and 0.58 over M0.
This suggests that even a fixed projection stabilises channel scales and token mixing at the selected layer.
Allowing this compact head to adapt at test time (M2) further increases accuracy to {68.29} and {77.91} mIoU, adding 1.96 and 2.13 over M1.
Cumulatively, M2 improves over M0 by 2.63 in one shot and 2.71 in 5-shot, which correspond to relative gains of about 4.0\% and 3.6\%, while keeping the fraction of updated parameters under 2.7\%.
These gains are consistent with the Select Regularise Calibrate design.
HLS provides a stable representation.
The small MLP recenters and rescales features to reduce support to query mismatch, and the resulting representations interact more reliably with PGR and PAC.
In practice, a single compact trainable MLP on the selected layer delivers most of the benefit with minimal overhead.

\section{Investigating the Effect of the Last Layer}
After HLS selects the best single-layer $\ell_{\text{single}}$ for each episode, we form a compact neighbourhood $U$ anchored at $\ell_{\text{single}}$ and we include the last ViT layer $L_{23}$ to mitigate fragmented shapes.
We evaluate all candidates under the same episodic objective.
For any $U$, let $r_\ell$ denote the single-layer ETR of layer $\ell$.
We compute the fusion weights and the fused representation as follows:
\begin{equation}
\begin{gathered}
w_\ell =
\frac{\exp\!\big(-\beta\, r_\ell - \mathrm{dist}(\ell, L_{23})/\tau\big)}
     {\sum_{j\in U}\exp\!\big(-\beta\, r_j - \mathrm{dist}(j, L_{23})/\tau\big)} \\[4pt]
F^{U} = \sum_{\ell\in U} w_\ell\, F^\ell
\end{gathered}
\label{eq:local-fusion}
\end{equation}
where $\beta>0$ controls reliance on the data evidence $r_\ell$, and $\tau>0$ is a locality bandwidth that biases the fusion toward deeper semantically aggregated layers.
As $\tau\!\to\!\infty$, the locality term vanishes and the solution reduces to single layer routing, that is $\arg\min_{\ell\in U} r_\ell$.
When evidence spreads across adjacent layers, a moderate $\tau$ balances data evidence and semantic aggregation and stabilises routing.

\begin{table}[h]
\centering
\footnotesize
\setlength{\tabcolsep}{2pt}
\caption{Local fusion anchored at the routed layer. We report average mIoU for the one-shot and five-shot settings, along with the changes relative to using $L_{23}$ alone and to excluding $L_{23}$.}
\label{tab:hlr_fusion_avg}
\begin{tabular}{l
S[table-format=2.2]
S[table-format=2.2]
S[table-format=+1.2]
S[table-format=2.2]
S[table-format=+1.2]}
\toprule
\multirow{2}{*}{Variant} & \multicolumn{2}{c}{mIoU avg.} & {\(\Delta\) vs. $L_{23}$} & {\(\Delta\) vs. no $L_{23}$} \\
\cmidrule(lr){2-3}\cmidrule(lr){4-4}\cmidrule(lr){5-5}
& {1 shot} & {5 shot} & {5 shot} & {5 shot} \\
\midrule
$F^0\!+\!L_{23}$, $\tau{=}0.0$       & 66.58 & 75.49 & 0.00 & +0.00 \\
$F^0\!+\!L_{23}$, $\tau{=}2.0$       & \bfseries 68.29 & \bfseries 77.85 & \bfseries +2.36 & \bfseries +2.36 \\
$F^0\!+\!\text{no }L_{23}$, pivot=last, $\tau{=}0.0$ & 66.45 & 75.29 & -0.20 & 0.00 \\
$F^0\!+\!\text{no }L_{23}$, pivot=$\ell^\star$, $\tau{=}2.0$ & 66.83 & 76.34 & +0.85 & +1.05 \\
\bottomrule
\end{tabular}
\end{table}

\begin{table*}[t]
\centering
\scriptsize
\setlength{\tabcolsep}{5pt}
\caption{By-dataset mIoU comparing $L_{23}$ alone and local fusion. Including $L_{23}$ in the candidate pool and setting $\tau{=}2.0$ yields the highest averages, with the largest gains on DeepGlobe and ISIC.}
\label{tab:hlr_fusion_dataset}
\resizebox{\textwidth}{!}{%
\begin{tabular}{lccccccccccc}
\toprule
\multirow{2}{*}{Backbone (DINOv3)} &
\multicolumn{2}{c}{DeepGlobe} &
\multicolumn{2}{c}{ISIC} &
\multicolumn{2}{c}{Chest X-ray} &
\multicolumn{2}{c}{FSS-1000} &
\multicolumn{2}{c}{Average} \\
\cmidrule(lr){2-3}\cmidrule(lr){4-5}\cmidrule(lr){6-7}\cmidrule(lr){8-9}\cmidrule(lr){10-11}
& 1-shot & 5-shot & 1-shot & 5-shot & 1-shot & 5-shot & 1-shot & 5-shot & 1-shot & 5-shot \\
\midrule
$F^0\!+\!L_{23}$, $\tau{=}2.0$ 
& \bfseries 44.59 & \bfseries 63.43 
& \bfseries 61.17 & \bfseries 73.64 
& 85.80 & 87.88 
& \bfseries 81.59 & \bfseries 86.69 
& \bfseries 68.29 & \bfseries 77.91 \\
$F^0\!+\!L_{23}$, $\tau{=}0.0$
& 42.90 & 61.49 
& 55.17 & 66.53 
& 87.06 & 88.29 
& 81.20 & 85.63 
& 66.58 & 75.49 \\
$F^0\!+\!\text{no }L_{23}$, pivot=last, $\tau{=}0.0$
& 42.87 & 61.43 
& 54.84 & 66.00 
& 87.01 & 88.26 
& 81.09 & 85.47 
& 66.45 & 75.29 \\
$F^0\!+\!\text{no }L_{23}$, pivot=$\ell^\star$, $\tau{=}2.0$
& 42.32 & 63.11 
& 56.41 & 68.25 
& \bfseries 87.44 & \bfseries 88.41 
& 81.16 & 85.58 
& 66.83 & 76.34 \\
\bottomrule
\end{tabular}}
\end{table*}

\vspace{-10pt}\paragraph{Analysis of L23.} Table~\ref{tab:hlr_fusion_avg} compares single-layer routing with local fusion. Local fusion anchored at $L_{23}$ with $\tau{=}2.0$ outperforms using $L_{23}$ alone on both one-shot and five-shot averages. Excluding $L_{23}$ from the candidate set reduces performance. Redirecting fusion to the routed layer $\ell^\star$ recovers part of the performance drop, yet it remains inferior to configurations that include $L_{23}$. By dataset, Table~\ref{tab:hlr_fusion_dataset} reports larger gains on DeepGlobe~\citep{demir2018deepglobe} and ISIC2018~\citep{tschandl2018ham10000,codella2019skin}, consistent with evidence drift across episodes and the need for deeper semantic aggregation. So we adopt local fusion with $\tau{=}2.0$ and retain $L_{23}$ in the candidate pool by default.

\section{Details of Pixelwise Adaptive Calibration}\label{app:pac}
Despite HLS and PGR, residual errors persist along thin boundaries, slender structures, and low contrast regions. With the backbone frozen, PAC adds three lightweight residual branches in the logit domain, coupled to the routed layer $\ell^\ast$ and to the patch attention calibrated by PGR.

\vspace{-8pt}\paragraph{Feature similarity for semantic alignment.}
Let $\mathbf{F}_q(x)$ denote the query feature at $\ell^\ast$. Foreground and background prototypes, $\mathbf{P}_{\mathrm{fg}}$ and $\mathbf{P}_{\mathrm{bg}}$, are computed by masked averaging over support features at $\ell^\ast$. We define the prototype difference logit as
\begin{equation}
\ell_{\mathrm{sim}}(x)
=\tau_{\mathrm{sim}}\!\left[
\cos\!\big(\mathbf{F}_q(x),\mathbf{P}_{\mathrm{fg}}\big)
-\cos\!\big(\mathbf{F}_q(x),\mathbf{P}_{\mathrm{bg}}\big)
\right],
\label{eq:pac-sim}
\end{equation}
where $\tau_{\mathrm{sim}}$ is a small temperature. This branch recovers missed regions and sharpens local focus.

\vspace{-8pt}\paragraph{One hop attention for spatial consistency.}
Let $\widetilde A$ denote the row-normalised patch to patch attention at $\ell^\ast$ after PGR. Given the base foreground probability $p_{0}(x)=\sigma(\ell_{0}(x))$, we propagate once on the patch grid as:
\begin{equation}
\ell_{\mathrm{attn}}(x)=\tau_{\mathrm{attn}}\!\big[(\widetilde A\,p_{0})_x\big],
\label{eq:pac-attn}
\end{equation}
This elongates responses along the object extent and suppresses spurious long-range peaks, with limited impact on the global distribution.

\vspace{-8pt}\paragraph{Image vector for appearance correction.}
Let $\mathbf{v}(x)$ denote a shallow appearance embedding for color and texture:
\begin{equation}
\ell_{\mathrm{img}}(x)
=\tau_{\mathrm{img}}\!\left[
\cos\!\big(\mathbf{v}(x),\mathbf{u}_{\mathrm{fg}}\big)
-\cos\!\big(\mathbf{v}(x),\mathbf{u}_{\mathrm{bg}}\big)
\right],
\label{eq:pac-img}
\end{equation}
Here $\mathbf{u}_{\mathrm{fg}}$ and $\mathbf{u}_{\mathrm{bg}}$ are image level prototypes, and $\tau_{\mathrm{img}}$ is a small temperature. This branch provides light global denoising and prevents over-shrinking.

\vspace{-8pt}\paragraph{The final logit} is a linear combination in the logit domain:
\begin{equation}
\ell_{\text{final}}(x)=
\ell_{0}(x)
+ w_{\text{sim}}\,\ell_{\text{sim}}(x)
+ w_{\text{attn}}\,\ell_{\text{attn}}(x)
+ w_{\text{img}}\,\ell_{\text{img}}(x),
\label{eq:pac-delta}
\end{equation}
where $\ell_{0}(x)$ is the base logit from the selected representation and $w_{\cdot}$ are fixed scalar weights. A single-step refine vote gate applies residuals only when the estimated gain is positive, adding negligible overhead. Together, the three stages realise a Select-Regularise-Calibrate pipeline, adapting at test time with a frozen backbone.

\section{Adaptive Behaviour of PAC}\label{app:gate}
After HLS and PGR, residual errors concentrate along thin boundaries and in low-contrast regions. Pixelwise Adaptive Calibration (PAC) adds three lightweight residual branches in the logit domain, namely feature similarity, one-hop attention propagation, and image appearance, while the backbone remains frozen.

\vspace{3pt}\noindent To avoid negative transfer, we enable PAC only when leave-one-out voting on the supports predicts a positive gain. Concretely, we treat each support as a pseudo query, compute the $\Delta\mathrm{mIoU}$ with and without PAC, and enable PAC on the true query if at least $T$ votes are positive. In the one-shot case, we synthesise two augmented views of the support to obtain two votes.

\begin{table}[h]
\centering
\small
\setlength{\tabcolsep}{10pt}
\caption{Effect of PAC gating thresholds. We report average mIoU (\%) and the trigger rate of the automatic gate. The best policy is to keep the gate always on for one shot, and to use automatic gating with a threshold $2/5$ for five shots.}
\label{tab:refine_threshold}
\begin{tabular}{lccc}
\toprule
Policy          & 1 shot & 5 shot & Trigger rate (auto) \\
\midrule
refine = off    & 67.54 & 76.67 & -- \\
auto, $T{=}1$   & 68.02 & --    & 56.32 \\
auto, $T{=}2$   & --    & \bfseries 77.91 & \bfseries 74.57 \\
auto, $T{=}3$   & --    & 77.22 & 59.44 \\
always on       & \bfseries 68.29 & 77.80 & -- \\
\bottomrule
\end{tabular}
\end{table}

\begin{table*}[h]
\centering
\small
\setlength{\tabcolsep}{4.5pt}
\caption{By-dataset mIoU and gate trigger rates. The recommended setting (one shot always on, five-shot automatic gating with threshold $2/5$) yields the highest average mIoU.}
\label{tab:refine_dataset}
\resizebox{\textwidth}{!}{%
\begin{tabular}{lcccccccccccc}
\toprule
& \multicolumn{2}{c}{DeepGlobe}
& \multicolumn{2}{c}{ISIC}
& \multicolumn{2}{c}{Chest X-ray}
& \multicolumn{2}{c}{FSS-1000}
& \multicolumn{2}{c}{Average}
& \multicolumn{2}{c}{Avg. trigger rate (\%)} \\
\cmidrule(lr){2-3}\cmidrule(lr){4-5}\cmidrule(lr){6-7}\cmidrule(lr){8-9}\cmidrule(lr){10-11}\cmidrule(lr){12-13}
Setting
& 1 shot & 5 shot
& 1 shot & 5 shot
& 1 shot & 5 shot
& 1 shot & 5 shot
& 1 shot & 5 shot
& 1 shot & 5 shot \\
\midrule
1-shot auto, 5-shot always
& 44.35 & 63.51
& 60.28 & 73.72
& 86.27 & 87.22
& 81.19 & 86.73
& 68.02 & 77.80
& \multicolumn{1}{c}{--} & \multicolumn{1}{c}{--} \\
Trigger rate (\%)
& \bfseries 55.83 & \multicolumn{1}{c}{--}
& \bfseries 50.00 & \multicolumn{1}{c}{--}
& \bfseries 19.50 & \multicolumn{1}{c}{--}
& \bfseries 99.95 & \multicolumn{1}{c}{--}
& \multicolumn{2}{c}{--}
& \bfseries 56.32 & \multicolumn{1}{c}{--} \\
\midrule
1 shot always, 5 shot auto 2/5
& 44.59 & 63.43
& 61.17 & 73.64
& 85.80 & 87.88
& 81.59 & 86.69
& \bfseries 68.29 & \bfseries 77.91
& \multicolumn{1}{c}{--} & \multicolumn{1}{c}{--} \\
Trigger rate (\%)
& \multicolumn{1}{c}{--} & \bfseries 25.67
& \multicolumn{1}{c}{--} & \bfseries 97.83
& \multicolumn{1}{c}{--} & \bfseries 25.67
& \multicolumn{1}{c}{--} & \bfseries 84.46
& \multicolumn{2}{c}{--}
& \multicolumn{1}{c}{--} & \bfseries 74.57 \\
\midrule
1 shot always, 5 shot auto 3/5
& 44.59 & 63.41
& 61.17 & 73.40
& 85.80 & 87.95
& 81.59 & 86.63
& \bfseries 68.29 & 77.85
& \multicolumn{1}{c}{--} & \multicolumn{1}{c}{--} \\
Trigger rate (\%)
& \multicolumn{1}{c}{--} & 5.83
& \multicolumn{1}{c}{--} & 69.00
& \multicolumn{1}{c}{--} & 5.83
& \multicolumn{1}{c}{--} & 74.92
& \multicolumn{2}{c}{--}
& \multicolumn{1}{c}{--} & 59.44 \\
\bottomrule
\end{tabular}}
\end{table*}

\vspace{-16pt}\paragraph{Analysis of gate.}
Relative to HLS at 76.7 mIoU, PGR raises the mean to 77.3 (+0.6), PAC to 77.2 (+0.5), and using PGR together with PAC yields 77.9 (+1.2), confirming their complementarity. For PAC gating, \cref{tab:refine_threshold} shows that in the one-shot setting, the best policy is to keep PAC on for all episodes (68.29 mIoU). In the five-shot setting, the automatic gate with threshold $T{=}2$ out of $5$ achieves the highest mean mIoU (77.91) with a moderate trigger rate (74.6\%), whereas $T{=}3$ out of $5$ further reduces the trigger rate but lowers accuracy to $77.22$–$77.85$ mIoU. The per-dataset study in \cref{tab:refine_dataset} supports the same recommendation: one shot with PAC on for all episodes and five-shot with automatic gating at $T{=}2$ out of $5$.

\vspace{3pt}\noindent Decomposing PAC on top of HLS plus PGR at 77.27 mIoU, the similarity residual $\ell_{\mathrm{sim}}$, the one-hop attention propagation $\ell_{\mathrm{attn}}$, and the image appearance cue $\ell_{\mathrm{img}}$ contribute +0.30, +0.22, and +0.18 mIoU, respectively. Using all three reaches 77.91 mIoU, a further +0.64. Together, HLS stabilises the routed layer, PGR sharpens locality, and PAC corrects pixel-level logits, yielding a cumulative gain under a frozen backbone.

\section{Episode-wise Layer Selectors}
\label{app:selectors}

\subsection{Episode notation and setting}
\label{sec:episode-notation}
Let $\ell\in\mathcal{C}$ index a ViT layer, and let $\mathbf{F}_q^\ell(x)\in\mathbb{R}^{d_\ell}$ denote the query feature at pixel $x$ from layer $\ell$. Support features are pooled using masks to form foreground and background prototypes $\mathbf{P}_{\mathrm{fg}}^\ell$ and $\mathbf{P}_{\mathrm{bg}}^\ell$. Given a baseline foreground probability $p_0(x)\in[0,1]$ for the query, we build soft masked query prototypes as:
\begin{equation}
\begin{gathered}
\mathbf{Q}_{\mathrm{fg}}^\ell
=\frac{\sum_x p_0(x)\,\mathbf{F}_q^\ell(x)}{\sum_x p_0(x)}, \\[3pt]
\mathbf{Q}_{\mathrm{bg}}^\ell
=\frac{\sum_x (1-p_0(x))\,\mathbf{F}_q^\ell(x)}{\sum_x (1-p_0(x))}.
\end{gathered}
\label{eq:qproto}
\end{equation}
Unless noted otherwise, all scalar layer scores are range normalised \emph{within each episode} across $\mathcal{C}$, so different selectors are comparable:
\begin{equation}
\begin{gathered}
\tilde s_\ell=\frac{s_\ell-\min_{j\in\mathcal{C}} s_j}{\max_{j\in\mathcal{C}} s_j-\min_{j\in\mathcal{C}} s_j+\operatorname{Var}epsilon},\\[3pt]
\qquad \operatorname{Var}epsilon=10^{-8}.
\end{gathered}
\end{equation}

\begin{table}[t]
\centering
\caption{Notation for layer selection in the episodic setting. All scalar layer scores are range-normalised across the candidate set $\mathcal{C}$ unless noted.}
\label{tab:notation-selectors}
\small
\begin{tabular}{@{}l l@{}}
\toprule
Symbol & Description \\
\midrule
$\ell\in\mathcal{C}$ & Candidate ViT layer index \\
$\mathbf{F}_q^\ell(x)\in\mathbb{R}^{d_\ell}$ & Query feature at pixel $x$ from layer $\ell$ \\
$\mathbf{P}_{\mathrm{fg}}^\ell,\ \mathbf{P}_{\mathrm{bg}}^\ell$ & Supp foreground/background prototypes at layer $\ell$ \\
$\mathbf{Q}_{\mathrm{fg}}^\ell,\ \mathbf{Q}_{\mathrm{bg}}^\ell$ & Soft masked query prototypes (see Eq.~\eqref{eq:qproto}) \\
$p_0(x)\in[0,1]$ & Baseline foreground probability on the query \\
$\mathrm{mIoU}_{\text{sup}}(\ell)$ & Support-only pseudo-query mIoU at layer $\ell$ \\
\bottomrule
\end{tabular}
\end{table}

\begin{table*}[t]
\centering
\footnotesize
\setlength{\tabcolsep}{4pt}
\caption{Layer selection ablation with DINOv2 (5-shot mIoU$\uparrow$). 
Rule lists the per episode selector and notation includes $\mathbf{g}_\ell$ gradient w.r.t.\ features of layer $\ell$; $S_{\text{sem}}$, $S_{\text{str}}$, $C$ normalized semantic, structure, and complexity scores; and $\mathrm{mIoU}_{\text{sup}}(\ell)$ support only mIoU.}
\label{tab:ablate_hlr_selection}
{\renewcommand{\arraystretch}{1.3}%
\begin{tabular*}{0.95\textwidth}{@{\extracolsep{\fill}}cc|C{1.6cm}C{1.6cm}C{1.6cm}C{1.6cm}|cc}
\toprule
Method & Rule & DeepGlobe & ISIC & Chest X\text{-}ray & FSS\text{-}1000 & Avg. & $\Delta$ \\
\midrule
Static-Max
& $\arg\max_{\ell}\big(\alpha\,S_{\text{sem}}(\ell)+\beta\,S_{\text{str}}(\ell)+\gamma\,C(\ell)\big)$
& \underline{50.8} & \underline{58.5} & \underline{79.0} & \underline{67.5} & \underline{64.0} & \underline{0.0} \\
Grad\text{-}Max 
& $\arg\max_{\ell}\ \lVert \mathbf{g}_{\ell}\rVert$
& 48.3 & 58.2 & 50.2 & 66.2 & 55.7 & -8.2 \\
Grad$\Delta$\text{-}Max 
& $\arg\max_{\ell}\ \big\lVert \mathbf{g}_{\ell}-\mathbf{g}_{\ell-1}\big\rVert$
& 48.5 & 58.3 & 50.1 & 66.3 & 55.8 & -8.2 \\
\textbf{HLS (ETR)} 
& $\arg\min_{\ell}\big(1-\mathrm{mIoU}_{\text{sup}}(\ell)\big)$
& \textbf{56.3} & \textbf{68.1} & \textbf{87.1} & \textbf{79.0} & \textbf{72.6}  & \textbf{+8.6} \\
\bottomrule
\end{tabular*}
}
\end{table*}

\subsection{Selectors other than HLS}
\label{sec:non-hls}
We group the non-episodic selectors into two families: a heuristic static rule built from prototype and mask scores, and gradient-based proxies. Unless noted, all scalar layer scores are range-normalised across the candidate set $\mathcal{C}$ within each episode. Prototypes and the baseline mask $p_0$ follow the definitions in \cref{sec:episode-notation}.

\vspace{-10pt}\paragraph{Static heuristic selector (Static-Max).}
This rule blends three normalised scores, namely semantic agreement, structure separation, and a complexity term combining texture and uncertainty, and selects the layer with the largest weighted sum:
\begin{equation}
\begin{gathered}
\ell^\star_{\text{static}}=
\operatorname*{arg\,max}_{\ell\in\mathcal{C}}
\big[\,\alpha' S_{\text{sem}}(\ell)+\beta' S_{\text{str}}(\ell)+\gamma' C(\ell)\,\big], \\[3pt] 
\alpha',\beta',\gamma'\!\ge\!0,\ \alpha'+\beta'+\gamma'=1,
\label{eq:static}
\end{gathered}
\end{equation}
where weights are task-specific, and the objective is a surrogate not directly tied to episode-level mIoU risk. And the Static-Max score is composed of three terms: a semantic agreement term, a structure separation term, and a texture–uncertainty complexity term.

\vspace{5pt}\noindent \textbf{Semantic agreement} measures how well the support and query prototypes align in feature space at layer $\ell$:
\begin{equation}
S_{\text{sem}}(\ell)=
\alpha\,\cos\!\big(\mathbf{P}_{\mathrm{fg}}^\ell,\mathbf{Q}_{\mathrm{fg}}^\ell\big)
+(1-\alpha)\,\cos\!\big(\mathbf{P}_{\mathrm{bg}}^\ell,\mathbf{Q}_{\mathrm{bg}}^\ell\big),
\label{eq:sem}
\end{equation}
where $\alpha\in[0,1]$ balances the contribution of foreground and background prototypes, $\mathbf{P}_{\mathrm{fg}}^\ell,\mathbf{P}_{\mathrm{bg}}^\ell$ are the support prototypes, and $\mathbf{Q}_{\mathrm{fg}}^\ell,\mathbf{Q}_{\mathrm{bg}}^\ell$ are the soft query prototypes at layer $\ell$. This term encourages higher agreement between support and query prototypes, but it depends on the baseline mask $p_0$, which can be biased under domain shift.

\vspace{5pt}\noindent \textbf{Structure separation} quantifies how well foreground and background features are separated in both the query and support spaces. It is defined as:
\begin{equation}
S_{\text{str}}(\ell)=1-\tfrac{1}{2}\!\left[
\cos\!\big(\mathbf{Q}_{\mathrm{fg}}^\ell,\mathbf{Q}_{\mathrm{bg}}^\ell\big)+
\cos\!\big(\mathbf{P}_{\mathrm{fg}}^\ell,\mathbf{P}_{\mathrm{bg}}^\ell\big)\right],
\label{eq:str}
\end{equation}
where the two cosine similarities measure the alignment between foreground and background features for the query and the supports, respectively. Large values of $S_{\text{str}}(\ell)$ indicate better foreground–background separation, but this term mainly captures feature-space geometry and is only indirectly related to the final mask quality.

\vspace{5pt}\noindent \textbf{Texture and uncertainty complexity} captures both the variability of foreground features and the ambiguity of the baseline mask. It is defined as:
\begin{equation}
\begin{gathered}
C(\ell) = \operatorname{Var}\!\big(\mathbf{Q}_{\mathrm{fg}}^\ell\big)
          + \operatorname{Ent}(p_0), \\[6pt]
\operatorname{Ent}(p_0)
= -\tfrac{1}{|\Omega|}\sum_{x}\!\big[
      p_0(x)\log p_0(x) \\[-2pt]
\qquad\qquad\quad
      + (1-p_0(x))\log\!\big(1-p_0(x)\big)
   \big],
\label{eq:comp}
\end{gathered}
\end{equation}
where $\operatorname{Var}(\cdot)$ denotes the per-dimension variance of query features relative to the corresponding foreground prototype, weighted by $p_0(x)$ over the pixel set $\Omega$, and $\operatorname{Ent}(p_0)$ is the average Bernoulli entropy of the baseline foreground probability. This term is an indirect proxy for semantic and appearance complexity and may penalise layers that are both confident and correct.

\paragraph{Gradient-based proxies.}
These rules, including Grad-Max and Grad$\Delta$-Max, preferentially select layers that show strong loss sensitivity or pronounced changes relative to neighbouring layers.

\vspace{5pt}\noindent\textbf{Gradient magnitude (Grad-Max)} measures how sensitive the base loss is to perturbations at layer $\ell$. It is defined as:
\begin{equation}
\ell^\star_{\text{grad}}=
\operatorname*{arg\,max}_{\ell\in\mathcal{C}}
\left\lVert\nabla_{\mathbf{F}_q^\ell}\mathcal{L}_{\text{base}}\right\rVert_2 ,
\label{eq:grad}
\end{equation}
where $\nabla_{\mathbf{F}_q^\ell}\mathcal{L}_{\text{base}}$ denotes the gradient of the base loss with respect to the query features at layer $\ell$, and the $\ell_2$-norm measures its magnitude. Larger values indicate stronger loss sensitivity at that layer.

\vspace{5pt}\noindent\textbf{Interlayer gradient change (Grad$\Delta$-Max)} measures sharp changes in gradient magnitude across adjacent layers. It is defined as:
\begin{equation}
\ell^\star_{\Delta\text{grad}}=
\operatorname*{arg\,max}_{\ell\in\mathcal{C}}
\left\lVert
\left\lVert\nabla_{\mathbf{F}_q^\ell}\mathcal{L}_{\text{base}}\right\rVert_2-
\left\lVert\nabla_{\mathbf{F}_q^{\ell-1}}\mathcal{L}_{\text{base}}\right\rVert_2
\right\rVert_2 ,
\label{eq:graddelta}
\end{equation}
where the difference of gradient magnitudes captures transitions between consecutive layers, and the outer $\ell_2$-norm measures the strength of this change.

\vspace{5pt}\noindent\textbf{Implementation notes.}
All gradient-based selectors reuse one forward pass of backbone features and require an additional backward pass \emph{without} parameter updates. The computational cost per episode is thus dominated by a single backpropagation through the frozen backbone.

\vspace{8pt}\subsection{Task-aligned HLS (ETR)}
\label{sec:hls-etr}
We select the routed layer by minimising an episode-level selection risk:
\begin{equation}
\begin{gathered}
R_{\text{layer}}(\ell)\;=\;1-miou_{\text{sup}}(\ell), \\[3pt]
\ell^{\star}_{\mathrm{HLS}} \;=\; \operatorname*{arg\,min}_{\ell\in\mathcal{C}} R_{\text{layer}}(\ell)
\;=\; \operatorname*{arg\,max}_{\ell\in\mathcal{C}}\space miou_{\text{sup}}(\ell),
\end{gathered}
\label{eq:hls}
\end{equation}
where $miou_{\text{sup}}(\ell)$ is computed within the episode by a leave-one-out procedure at layer $\ell$. Each support image is treated as a pseudo query and segmented using prototypes formed from the remaining supports, and the result is averaged over the $K$ supports.

\vspace{5pt}\noindent The criterion in Eq.~\eqref{eq:hls} directly measures episode-level matching risk at the representation to be adapted, rather than optimising a handcrafted surrogate. This makes it robust to layer-level transfer variability and domain shift. In practice, HLS is parameter-free, reuses the same forward features, and adds negligible overhead.

\begin{table*}[!t]
\centering
\footnotesize
\setlength{\tabcolsep}{3pt}
\caption{Generalisation of HERA across different ViT-based VFMs on the CD-FSS benchmark.
``ONLY'' denotes directly using the frozen VFM within SSP, while ``HERA'' adds our three-stage
select-regularise-calibrate VFM-based segmentation framework. The last two columns report the
average mIoU gains of HERA over the corresponding VFM-ONLY baselines.}
\resizebox{\linewidth}{!}{%
\begin{tabular}{@{} c | cc cc cc cc cc G G @{}}
\toprule
\multirow{2}{*}{Backbone} &
\multicolumn{2}{c}{DeepGlobe} &
\multicolumn{2}{c}{ISIC2018} &
\multicolumn{2}{c}{Chest X-ray} &
\multicolumn{2}{c}{FSS-1000} &
\multicolumn{2}{c}{Average} &
\multicolumn{2}{c}{Gain w/ HERA} \\
\cmidrule(lr){2-3}\cmidrule(lr){4-5}\cmidrule(lr){6-7}\cmidrule(lr){8-9}\cmidrule(lr){10-11}\cmidrule(lr){12-13}
& 1-shot & 5-shot & 1-shot & 5-shot & 1-shot & 5-shot & 1-shot & 5-shot & 1-shot & 5-shot & 1-shot & 5-shot \\
\midrule
SSP                           & 40.5 & 49.6 & 35.5 & 48.2 & 74.2 & 74.5 & \secbest{79.0} & 80.2 & 57.3 & 63.1 & 0.0       & 0.0       \\
\midrule
DINOv2-ONLY (Large)           & 43.9 & 52.9 & 50.7 & 56.6 & 51.2 & 49.1 & 67.4 & 67.9 & 53.3 & 56.6 &        &        \\
DINOv2-HERA (Large)           & 41.2 & 57.8 & 55.6 & \secbest{68.7} & \secbest{83.2} & \secbest{86.9} & 70.2 & \secbest{80.3} & \secbest{62.6} & \secbest{73.4} & +9.3 & \secbest{+16.8} \\
\midrule
DINOv3-ONLY (Large)           & \best{47.1} & \secbest{58.4} & \secbest{56.3} & 61.7 & 56.4 & 60.2 & 75.8 & 76.4 & 58.9 & 64.2 &        &        \\
DINOv3-HERA (Large)           & \secbest{44.6} & \best{63.4} &
                                \best{61.2} & \best{73.6} &
                                \best{85.8} & \best{87.9} &
                                \best{81.6} & \best{86.7} &
                                \best{68.3} & \best{77.9} &
                                \secbest{+9.4} & +13.7 \\
\midrule
CLIP-L/14-ONLY (OpenAI)       & 20.4 & 24.4 & 30.1 & 37.4 & 48.6 & 48.4 & 53.8 & 54.5 & 38.2 & 41.2 &        &        \\
CLIP-L/14-HERA (OpenAI)       & 24.3 & 38.1 & 37.8 & 53.0 & 68.4 & 82.2 & 61.8 & 74.2 & 48.1 & 61.9 & \best{+9.9} & \best{+20.7} \\
\midrule
CLIP-L/14-ONLY (DataComp-XL)  & 23.3 & 27.5 & 39.9 & 46.2 & 59.2 & 59.4 & 58.4 & 59.5 & 45.2 & 48.2 &        &        \\
CLIP-L/14-HERA (DataComp-XL)  & 27.2 & 38.5 & 40.1 & 53.1 & 78.1 & 83.8 & 66.9 & 73.4 & 53.1 & 62.2 & +7.9 & +14.0 \\
\bottomrule
\end{tabular}%
}
\label{tab:vfm_generalisation}
\end{table*}

\vspace{8pt}\subsection{Selector analysis and takeaway}
\label{sec:selector-analysis}

\textbf{Limitations of the non-episodic selectors.} We compare per-episode selectors for test-time adaptation with a frozen backbone, as shown in Table~\ref{tab:ablate_hlr_selection}. The \emph{Static-Max} rule blends three normalized cues and selects the layer with the largest $\alpha' S_{\text{sem}}(\ell)+\beta' S_{\text{str}}(\ell)+\gamma' C(\ell)$ (see Eqs.~\eqref{eq:sem} to \eqref{eq:comp}). These scores measure representation quality in feature space, including semantic agreement, structure separation, and texture or uncertainty, but they do not measure \emph{task fit} for the episode. They lack episode-level feedback and are therefore unstable across domains. Specifically, $S_{\text{sem}}$ inherits bias from the baseline mask $p_0$, $S_{\text{str}}$ rewards orthogonality that does not guarantee correct masks, and $C(\ell)$ can penalise layers that are confident and correct. The mixture weights $\alpha',\beta',\gamma'$ are domain-specific. Consequently, Static-Max averages 64.0 mIoU.

\vspace{5pt}\noindent Gradient-based proxies capture loss sensitivity rather than alignment. \emph{Grad-Max} selects the layer with the largest gradient norm (see~\cref{eq:grad}), and \emph{Grad$\Delta$-Max} looks for sharp inter-layer gradient changes (see~\cref{eq:graddelta}). In ViT backbones such as \textsc{DINOv2~\citep{oquab2023dinov2}} and \textsc{DINOv3~\citep{simeoni2025dinov3}}, blocks are architecturally homogeneous and connected by residual paths and layer normalisation. This can cause gradients to grow toward the last blocks, so both rules tend to collapse to deep layers irrespective of the episode semantics. This Grad-CAM-style assumption therefore fails, and the selected layer often has the largest perturbation rather than being the most suitable for segmentation. These proxies correlate weakly with support and query matching quality and yield 55.7 and 55.8 mIoU on average.

\vspace{5pt}\noindent\textbf{Why HLS (ETR) is better.} 
Our \emph{HLS} uses a task-aligned criterion that directly minimises the episode-level selection risk $\ell^\star_{\mathrm{HLS}}=\arg\min_{\ell\in\mathcal{C}}\big(1-miou_{\text{sup}}(\ell)\big)$ (see Eq.~\eqref{eq:hls}). It performs a self-prediction evaluation within the episode. Each support is treated as a pseudo query and is segmented using prototypes from the remaining supports, and the score is the support-only mIoU at layer $\ell$. This provides dynamic, episode-aware feedback aligned with the target objective, with no extra parameters, and negligible overhead. HLS reaches 72.6 mIoU, which is +8.6 over Static-Max and +16.8 over the best gradient proxy. The gain is especially large on Chest X-ray~\citep{candemir2013lung,jaeger2013automatic} (from 50.1 to 87.1 mIoU, +37.0), and the gap widens on other VFM backbones.

\section{Generalising HERA across ViT-based VFMs}
\label{sec:vfm_generalisation}

In this section, we evaluate the generalisation of HERA across different ViT-based visual foundation models (VFMs). We use SSP as a common few-shot segmentation framework and treat several representative ViT-based VFMs as frozen backbones. Specifically, we consider the self-supervised \textsc{DINOv2}~\citep{oquab2023dinov2} and \textsc{DINOv3}~\citep{simeoni2025dinov3}, and the contrastively trained CLIP-L/14 (OpenAI)~\citep{radford2021learning} and CLIP-L/14 (DataComp-XL)~\citep{gadre2023datacomp}. On the CD-FSS benchmark, we keep all training and evaluation settings identical to those in the main experiments. The only degrees of freedom are the choice of backbone and whether HERA is applied. In Table~\ref{tab:vfm_generalisation}, the ``ONLY'' rows correspond to directly using the frozen VFM within SSP, whereas the ``HERA'' rows apply our three-stage select--regularise--calibrate adaptation framework on top of the same VFM. The last two columns report the average mIoU gains achieved by HERA over the corresponding VFM-ONLY baselines.

\vspace{5pt}\noindent
HERA yields consistent and substantial relative improvements across the ViT-based VFMs. For self-supervised VFMs, DINOv2-HERA increases the average mIoU from 53.3/56.6 to 62.6/73.4, corresponding to gains of \textbf{+9.3} and \textbf{+16.8} in the 1-shot and 5-shot settings, respectively. DINOv3-HERA improves over DINOv3-ONLY from 58.9/64.2 to 68.3/77.9, corresponding to \textbf{+9.4} and \textbf{+13.7} mIoU, and surpasses 80\% mIoU on several targets. On CLIP-based VFMs, HERA still shows clear and consistent gains. For OpenAI CLIP-L/14, the average mIoU increases from 38.2/41.2 to 48.1/61.9, i.e\onedot\, gains of \textbf{+9.9} and \textbf{+20.7}. For DataComp-XL CLIP-L/14, HERA delivers improvements of \textbf{+7.9} and \textbf{+14.0} in the 1-shot and 5-shot regimes, respectively. These results demonstrate that, under a unified hyperparameter setting and without any source-domain retraining, HERA systematically enhances the performance of diverse ViT-based VFMs on CD-FSS.

\vspace{5pt}\noindent
In  CD-FSS, CLIP is primarily pre-trained for image-text alignment and image-level recognition, so its patch-level representations are naturally more global and less tailored to precise segmentation boundaries. This mismatch makes CD-FSS a particularly challenging downstream task for CLIP. Nevertheless, HERA even achieves some of the largest relative gains on CLIP backbones, especially in the 5-shot setting. This indicates that our select--regularise--calibrate mechanism effectively mitigates the mismatch between pre-training objectives and dense prediction. When the underlying VFM is more aligned with dense recognition, as in the case of DINOv3, HERA further amplifies its potential and achieves strong, often state-of-the-art performance on CD-FSS. Overall, these observations suggest that HERA can serve as a generic adaptor for ViT-based VFMs and is well-positioned to benefit future VFMs that are even better suited to class-disjoint few-shot segmentation.

\section{Component Ablation of DINOv2}
\label{app:dinov2-ablation}
The sequence Select-Regularise-Calibrate yields monotonic improvements.
\text{HLS} provides the dominant gain by stabilising the chosen adaptation layer for each episode.
\text{PGR} reduces attention noise, such as spurious far-field peaks, while preserving global coverage.
\text{PAC} then corrects residual artifacts along thin boundaries and in low-contrast regions.
Gains are larger in the one-shot regime, where supervision is scarcer, which is consistent with the design intent.
These results show that the hierarchical refinements generalise from DINOv3~\citep{simeoni2025dinov3} to DINOv2~\citep{oquab2023dinov2} and to other VFMs, indicating effectiveness that is agnostic to the backbone.

\begin{table}[h]
\centering
\scriptsize
\setlength{\tabcolsep}{6pt}
\caption{Component ablation on DINOv2 (average mIoU). $\Delta_{\mathrm{V0}}$ denotes the improvement over the V0 baseline, and $\Delta_{\mathrm{prev}}$ denotes the improvement relative to the row above. Best scores in bold.}
\label{tab:appendix_dinov2_ablate}
\begin{tabular}{lcccc}
\toprule
Setting & Avg. 1-shot & Avg. 5-shot & $\Delta_{\text{V0}}$ (1s / 5s) & $\Delta_{\text{prev}}$ (1s / 5s) \\
\midrule
V0 baseline  & 57.03 & 68.49 & 0.00 / 0.00 & 0.00 / 0.00 \\
+ HLS           & 60.34 & 72.64 & +3.31 / +4.15 & \textbf{+3.31} / \textbf{+4.15} \\
+ PGR       & 61.10 & 73.28 & +4.07 / +4.79 & +0.76 / +0.64 \\
+ PAC                        & \bfseries 62.58 & \bfseries 73.42 & \bfseries +5.55 / +4.93 & +1.48 / +0.14 \\
\bottomrule
\end{tabular}
\end{table}

\vspace{5pt}\noindent All selectors reuse cached features.
HLS uses pseudo-query scoring on the support only and therefore adds negligible overhead.
PGR has no trainable parameters.
PAC operates as a lightweight residual fusion and is gated automatically in five-shot episodes.
Consequently, the overall parameter and runtime budgets remain low while providing improvements that are aligned with the task.

\section{1-shot Support Augmentation Study}
\label{app:1shot_aug}

In the 1-shot setting, we explore a lightweight support-augmentation strategy designed to stabilise the leave-one-out scoring used by HLS. To maintain compatibility with HERA’s prototype-construction and refinement pipeline, we adopt a simple \emph{soft-copy} instance-synthesis procedure. Given a support image and its foreground mask, we paste the foreground onto a randomly sampled location within the same image to generate additional pseudo-supports while maintaining mask consistency. Formally, for a foreground instance $(I_{\mathrm{fg}}, M_{\mathrm{fg}})$ and a background view $(I_{\mathrm{bg}}, M_{\mathrm{bg}})$, the synthesized example is:
\begin{equation}
\begin{aligned}
I'_{\mathrm{bg}} &= I_{\mathrm{bg}}
+ M_{\mathrm{fg}}^{(x,y)} \odot I_{\mathrm{fg}}, \\[2pt]
M'_{\mathrm{bg}} &= \operatorname{clip}\!\big(
M_{\mathrm{bg}} + M_{\mathrm{fg}}^{(x,y)},\, 0,1\big),
\end{aligned}
\label{eq:softcopy}
\end{equation}
where $M_{\mathrm{fg}}^{(x,y)}$ denotes the foreground mask placed at a random location $(x,y)$, and $\operatorname{clip}$ ensures that the resulting mask remains within $[0,1]$. This soft-copy operation preserves foreground semantics while introducing controlled variations in appearance.

\vspace{5pt}\noindent Across backbones and datasets, we have an observation about soft-copy augmentation: the synthesised pseudo-supports remain highly correlated in semantic structure, causing the leave-one-out estimates used by HLS to fluctuate only minimally. Nevertheless, HERA already maintains competitive 1-shot performance under this simple augmentation mechanism.

\vspace{5pt}\noindent Importantly, this observation shows that our framework is highly compatible with more powerful support-augmentation strategies. Because HERA only requires soft masks, foreground prototypes, and within-episode consistency measures, it can seamlessly incorporate more advanced augmentations, including generative instance synthesis, structure-preserving shape deformation, and domain-aware appearance perturbations. These directions offer substantial potential for improving 1-shot stability and overall cross-domain generalisation capability.

\vspace{5pt}\noindent In summary, soft-copy serves as a minimal yet stable augmentation method for 1-shot episodes, demonstrating that HERA remains competitive even under limited synthetic diversity. More expressive support-synthesis strategies remain compatible with our design and provide promising avenues for further improving robust and scalable 1-shot performance across diverse domains.

\section{More Visualisation Results about HERA}
We provide additional qualitative results under the 1-shot setting to support the analyses in this paper. We include (i) segmentation predictions across four target datasets and (ii) layer-wise foreground probability maps from all ViT layers, which visually illustrate the layer sensitivity phenomenon and the semantic transition exploited by HERA.
\begin{figure}[h]
  \centering
  \includegraphics[page=1,width=\linewidth]{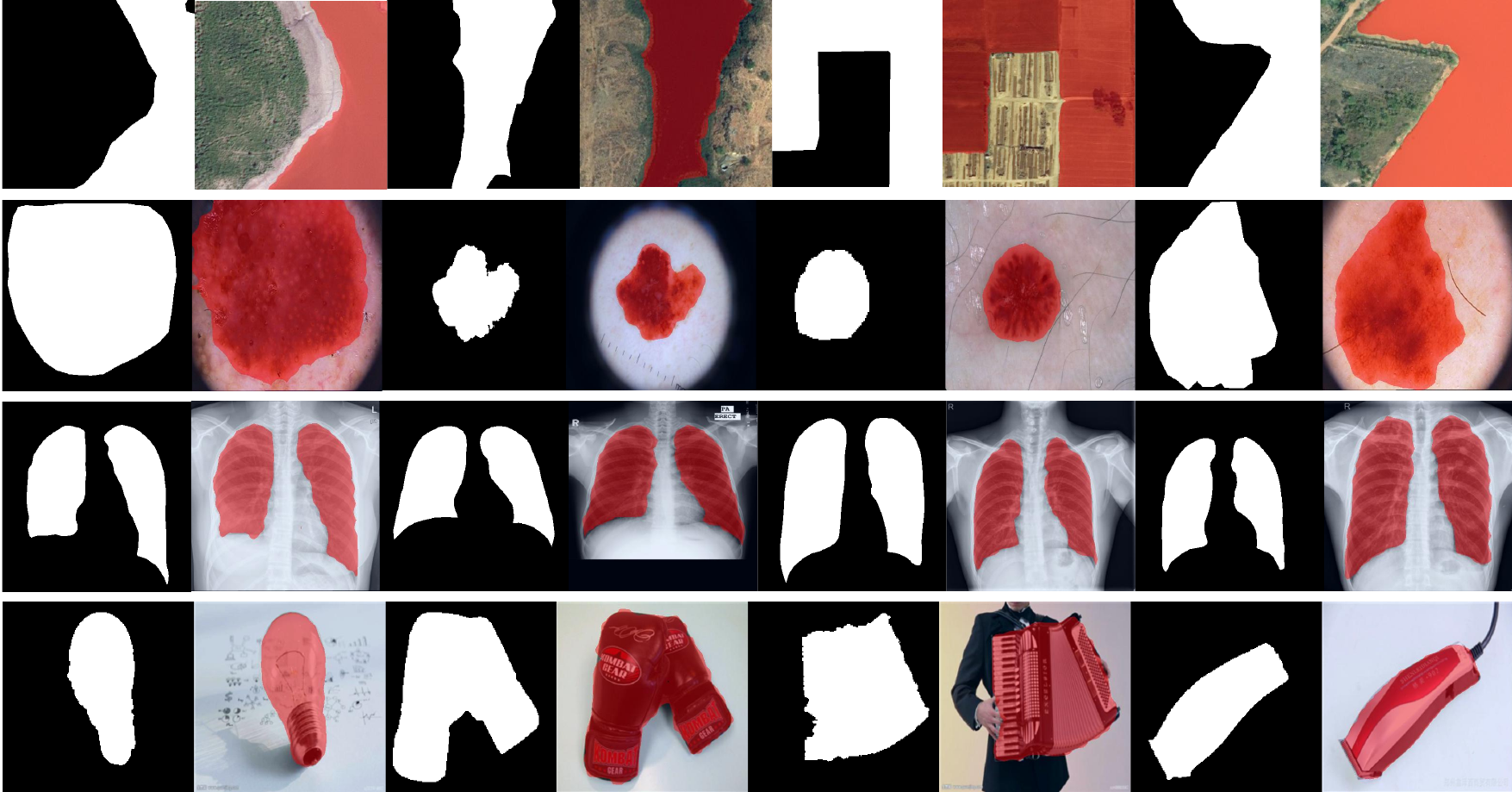}
  \caption{\small 1-shot segmentation results on the DeepGlobe, ISIC, Chest X-ray and FSS-1000 datasets, showing the predicted masks overlaid on the query images.}
  \label{fig:Qualitative}
  \vspace{-0.5em}
\end{figure}

\begin{figure}[h]
  \centering
  \includegraphics[page=1,width=\linewidth]{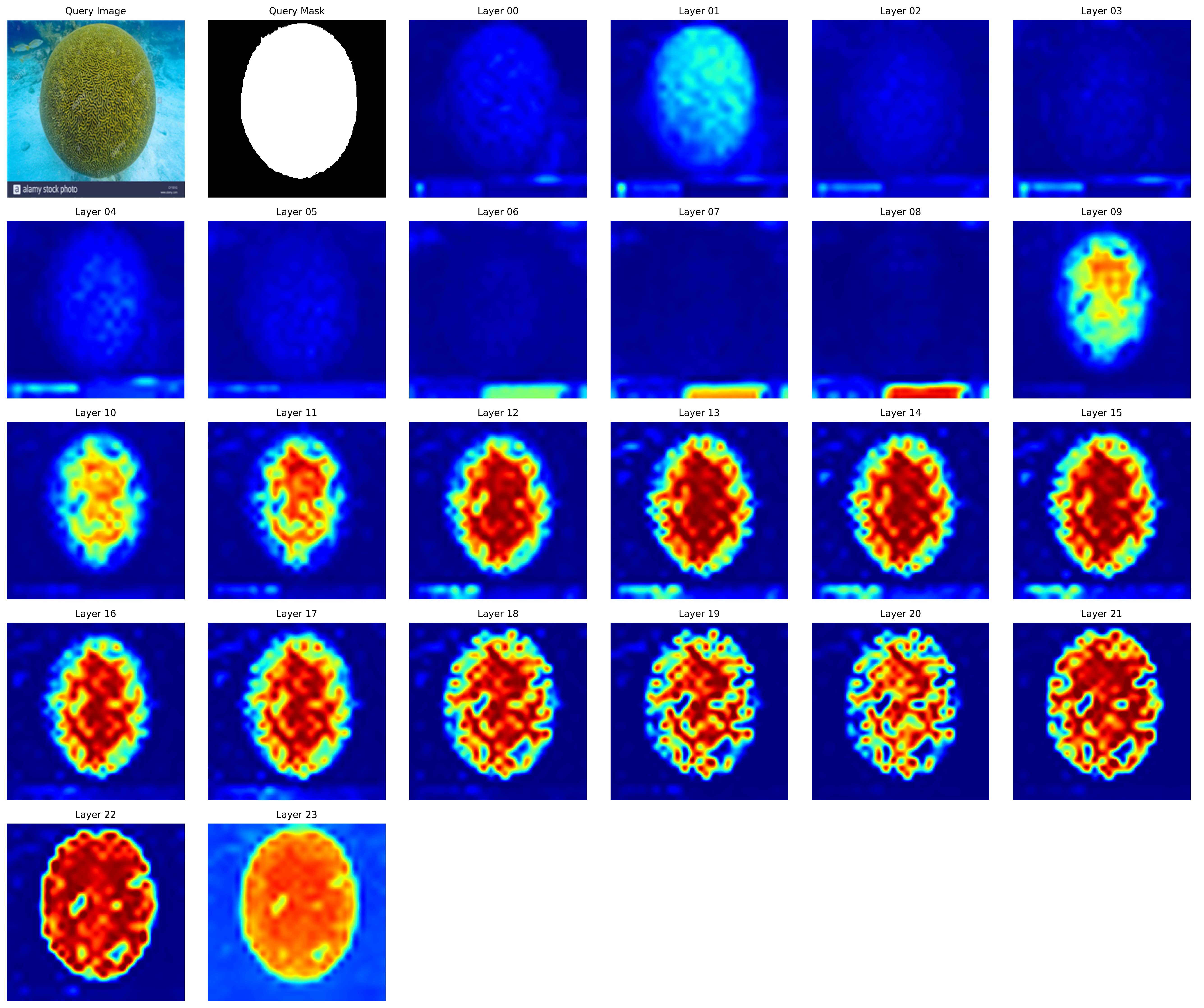}
  \caption{\small Layer-wise predicted foreground probability maps (Layer 00–23) showing the emergence of semantic structures across depth under the 1-shot setting.}
  \label{fig:Qualitative}
  \vspace{-0.5em}
\end{figure}

\begin{figure}[h]
  \centering
  \includegraphics[page=1,width=\linewidth]{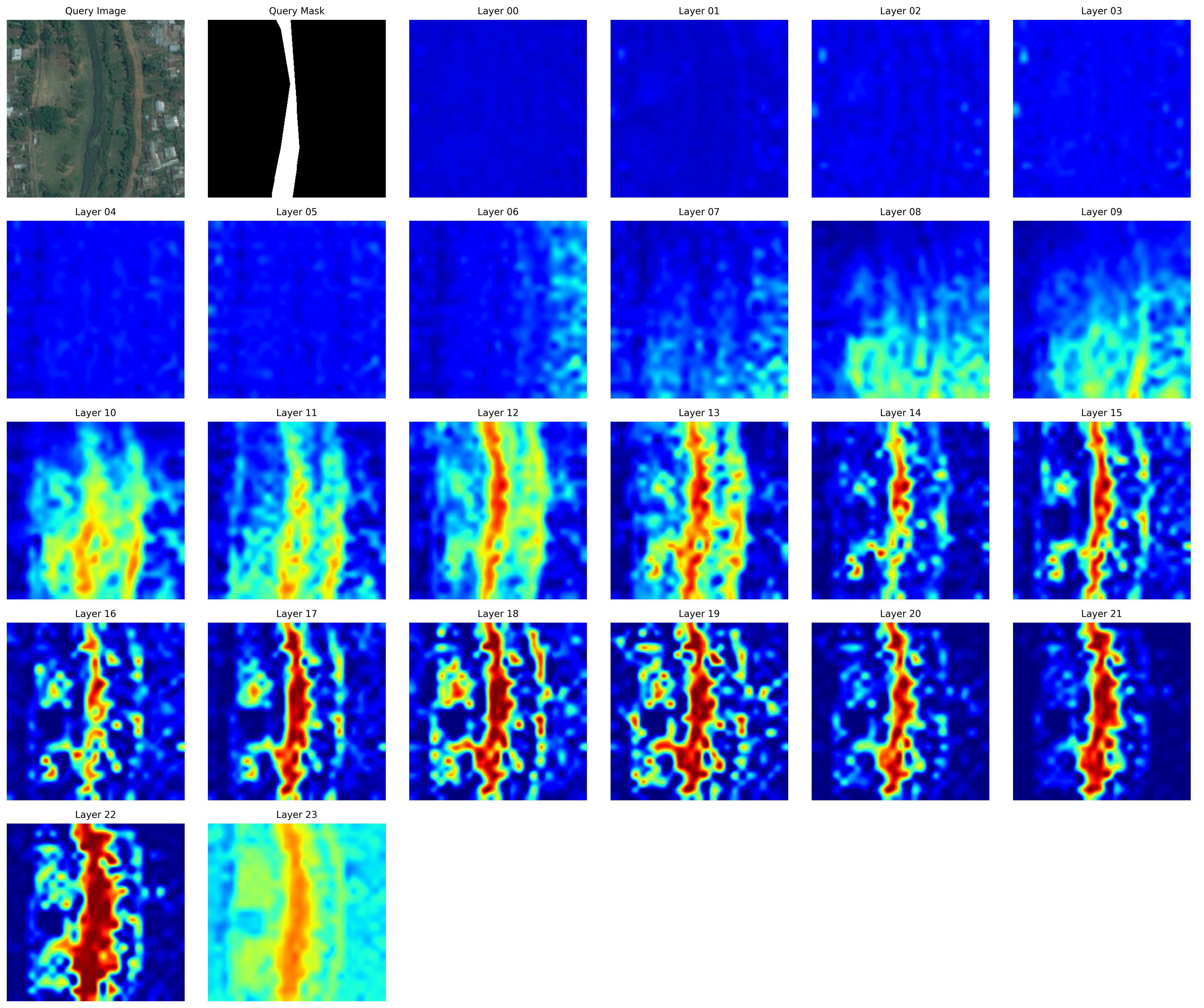}
  \caption{\small Additional layer-wise probability maps demonstrating consistent semantic transitions across diverse 1-shot episodes.}
  \label{fig:Qualitative}
  \vspace{-0.5em}
\end{figure}

\begin{figure}[t]
  \centering
  \includegraphics[page=1,width=0.8\linewidth]{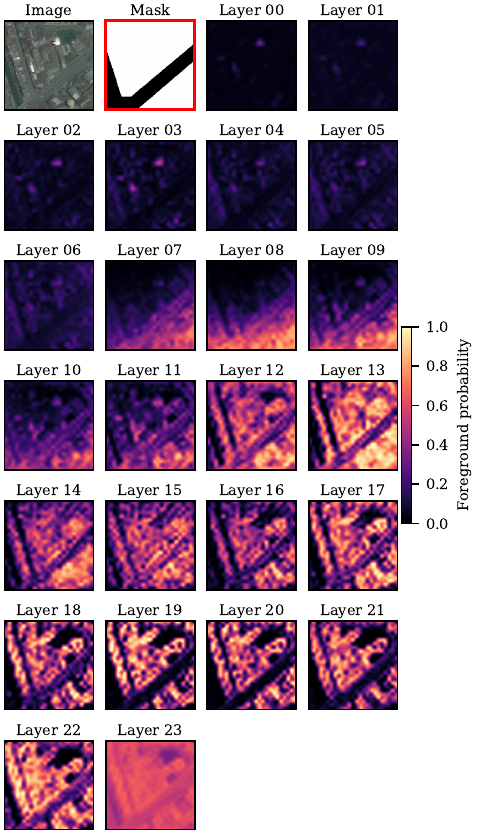}
  \caption{\small Per-layer probability responses for a representative 1-shot query, highlighting the mid-layer semantic turning point captured by HERA.}
  \label{fig:Qualitative}
  \vspace{-0.5em}
\end{figure}

\begin{figure}[t]
  \centering
  \includegraphics[page=1,width=\linewidth]{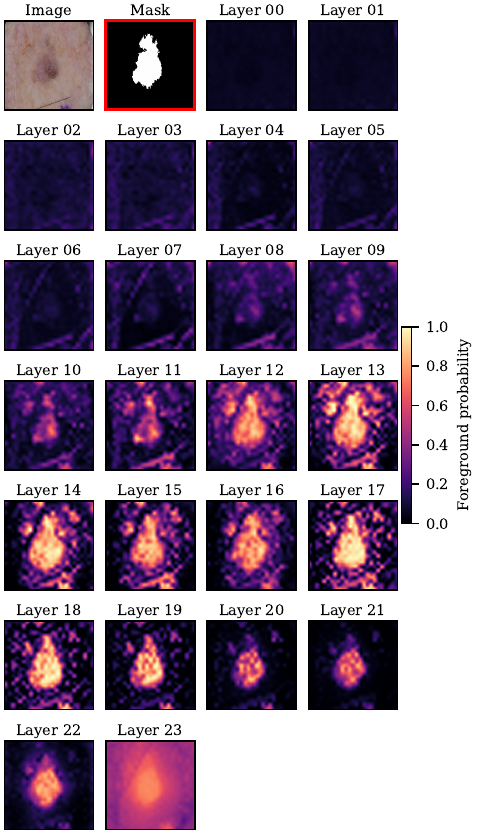}
  \caption{\small Layer-wise probability visualisation for another 1-shot sample, illustrating stable depth-dependent semantic strengthening across datasets.}
  \label{fig:Qualitative}
  \vspace{-0.5em}
\end{figure}

\begin{figure}[t]
  \centering
  \includegraphics[page=1,width=\linewidth]{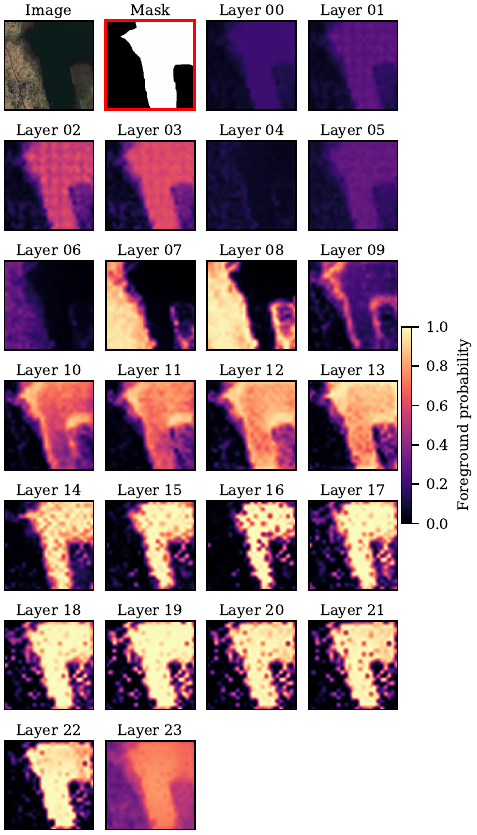}
  \caption{\small Foreground probability maps across all ViT layers for a third 1-shot episode, revealing the deep-layer aggregation pattern leveraged by the proposed routing mechanism.}
  \label{fig:Qualitative}
  \vspace{-0.5em}
\end{figure}